\documentclass[lettersize,journal]{IEEEtran}
\usepackage{amsmath,amsfonts}
\usepackage{array}
\usepackage[caption=false,font=normalsize,labelfont=sf,textfont=sf]{subfig}
\usepackage{textcomp}
\usepackage{stfloats}
\usepackage{url}
\usepackage{verbatim}
\usepackage{graphicx}
\usepackage{cite}

\usepackage{enumitem}

\usepackage{multirow}

\usepackage{graphicx} 

\usepackage[labelfont=bf]{caption}

\usepackage{color}

\usepackage{tabu}

\usepackage{booktabs}

\usepackage{array}
\usepackage{url} 
\usepackage{algorithmicx}
\usepackage{tabularx,booktabs}
\usepackage{graphicx}
\makeatletter
\newif\if@restonecol
\makeatother

\usepackage[linesnumbered,ruled,vlined]{algorithm2e}
\usepackage{algpseudocode}
\usepackage{amsmath}
\newcolumntype{C}[1]{>{\centering\let\newline\\\arraybackslash\hspace{0pt}}m{#1}}
\usepackage{tabularx} 
\newcolumntype{C}{>{\centering\arraybackslash}X} 
\begin{document}
\newcommand{\aliasAPP}{OCTOPUS\xspace}
\title{Exploring Deep Reinforcement Learning for Holistic Smart Building Control}

\author{Xianzhong~Ding,~\IEEEmembership{}
        Alberto~Cerpa,~\IEEEmembership{Member,~IEEE,}
        and~Wan~Du,~\IEEEmembership{Member,~IEEE}
\thanks{A preliminary version of this work was published in the Proceedings of ACM BuildSys 2019 \cite{ding2019octopus}.}


}



\maketitle

\begin{abstract}
Recently, significant efforts have been done to improve quality of comfort for commercial buildings' users while also trying to reduce energy use and costs. Most of these efforts have concentrated in energy efficient control of the HVAC (Heating, Ventilation, and Air conditioning) system, which is usually the core system in charge of controlling buildings' conditioning and ventilation.  However, in practice, HVAC systems alone cannot control every aspect of conditioning and comfort that affects buildings' occupants. Modern lighting, blind and window systems, usually considered as independent systems, when present, can significantly affect building energy use, and perhaps more importantly, user comfort in terms of thermal, air quality and illumination conditions. For example, it has been shown that a blind system can provide 12\%$\sim$35\% reduction in cooling load in summer while also improving visual comfort. In this paper, we take a holistic approach to deal with the trade-offs between energy use and comfort in commercial buildings. We developed a system called \aliasAPP, which employs a novel deep reinforcement learning (DRL) framework that uses a data-driven approach to find the optimal control sequences of \emph{all} building's subsystems, including HVAC, lighting, blind and window systems.  The DRL architecture includes a novel reward function that allows the framework to explore the trade-offs between energy use and users' comfort, while at the same time enable the solution of the high-dimensional control problem due to the interactions of four different building subsystems. In order to cope with \aliasAPP's data training requirements, we argue that calibrated simulations that match the target building operational points are the vehicle to generate enough data to be able to train our DRL framework to find the control solution for the target building.  In our work, we trained \aliasAPP with 10-year weather data and a building model that is implemented in the EnergyPlus building simulator, which was calibrated using data from a real production building. Through extensive simulations we demonstrate that \aliasAPP can achieve 14.26\% and 8.1\% energy savings compared with the state-of-the art rule-based method in a LEED Gold Certified building and the latest DRL-based method available in the literature respectively, while maintaining human comfort within a desired range.
\end{abstract}


\section{Introduction}
Energy saving in buildings is important to society, as buildings consume 32\% energy and 51\% electricity demand worldwide~\cite{lucon2014buildings, rajabi2022modes}.
Rule-based control (RBC) is widely used to set the actuators (e.g., heating or cooling temperature, and fan speed) in the HVAC (heating, ventilation, and air-conditioning) system. 
The "rules" in RBC are usually set as some static thresholds or simple control loops based on the experience of engineers and facility managers. 
The thresholds and simple control rules may not be optimal and have to be adapted to new buildings at commissioning time. Many times these rules are updated in an ad-hoc manner, based on experience and feedback from occupants and/or trial and error performed by HVAC engineers during the operational use of the building.
As a result, many model-based approaches have been developed to model the thermal dynamics of a building and execute a control algorithm on top of the model, such as Proportional Integral Derivative (PID)~\cite{shein2012pid} and Model Predictive Control (MPC)~\cite{beltran2014optimal}. However, the complexity of the thermal dynamics and the various influencing factors are hard to be precisely modeled, which is why the models tend to be simplified in order deal with the parameter-fitting data requirements and computational complexity when solving the optimization problem~\cite{beltran2014optimal}.

To tackle the limitations of the model-based methods, some model-free approaches have been proposed based on reinforcement learning (RL) for HVAC control, including Q-learning~\cite{li2015multi} and Deep Reinforcement Learning (DRL)~\cite{zhang2018practical}. 
With RL, an optimal control policy can be learned by the trial-and-error interaction between a control agent and a building, without explicitly modeling the system dynamics. 
By adopting a deep neural network as the control agent, DRL-based schemes can handle large state and action space in building control \cite{zhang2018practical}.
Some recent work~\cite{wei2017deep, zhang2018practical} has shown that DRL can provide real-time control for building energy efficiency.
However, all existing methods only consider a single subsystem in buildings, e.g., the HVAC system~\cite{wei2017deep} or the heating system~\cite{zhang2018practical}, ignoring some other subsystems that can affect performance from the energy use and/or user comfort point of view.

At present, more and more buildings are been equipped with automatically-adjustable windows and blinds. 
For example, motor-operated windows and blinds, like the intelligent products from GEZE~\cite{smart_window}, have been installed using an effective natural ventilation strategy~\cite{smart_window_article}.
In addition, researchers have studied the potential of energy saving by jointly controlling the HVAC system and another subsystem, like blind~\cite{tzempelikos2007impact}, lighting~\cite{cheng2016satisfaction}, and window~\cite{wang2015window}. 
For example, the energy consumed by HVAC can be reduced by 17\%$\sim$47\% if window-based natural ventilation is enabled~\cite{wang2015window}. 

\begin{figure}[t]
\centering
  \includegraphics[height=1.6in, width=3.1in]{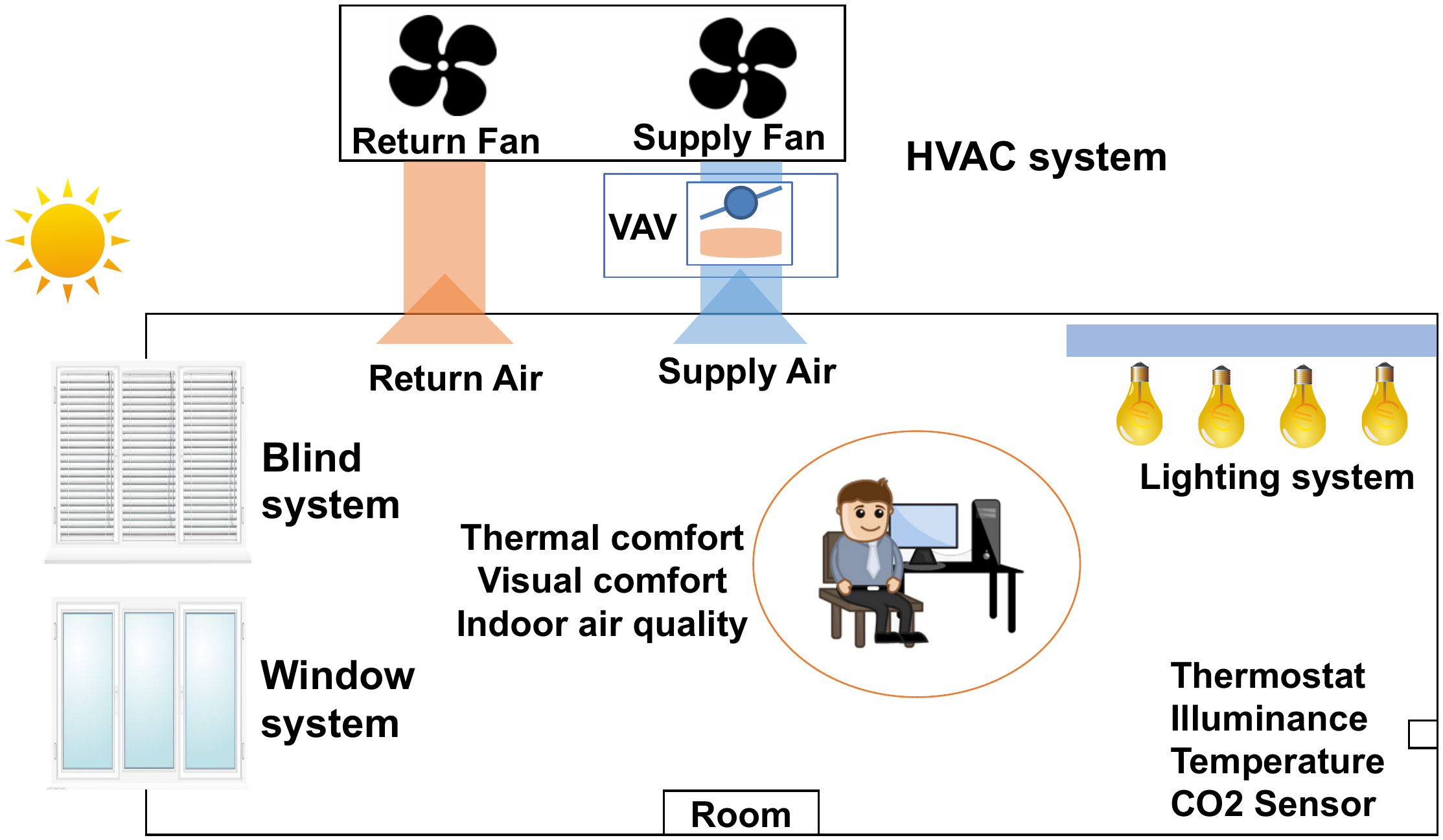}
  \caption{Four Subsystems in a Typical Building.}
  \label{combined_system}
\end{figure}

In this work, we argue that a \emph{holistic approach} that considers \emph{all available subsystems} (HVAC, blinds, windows, lights) in buildings, which have complex and non-trivial interactions should be used in coordination to achieve a specific energy efficiency/comfort goal.
Figure~\ref{combined_system} shows a depiction of a modern building that includes multiple subsystems (e.g., HVAC, window, blind and lighting) that work together to guarantee human comfort goals, including thermal comfort, visual comfort, and indoor air quality goals.
For example, indoor temperature can be influenced by three subsystems, like setting the HVAC temperature (adjusting the discharge temperature set points at the VAV level), and/or adjusting blind slats (allowing external sunlight to heat indoor air) and/or the window system (enabling exchange of indoor and outdoor air).

\begin{figure}[t]
\begin{center}
  \includegraphics[height=0.6in, width=2.8in]{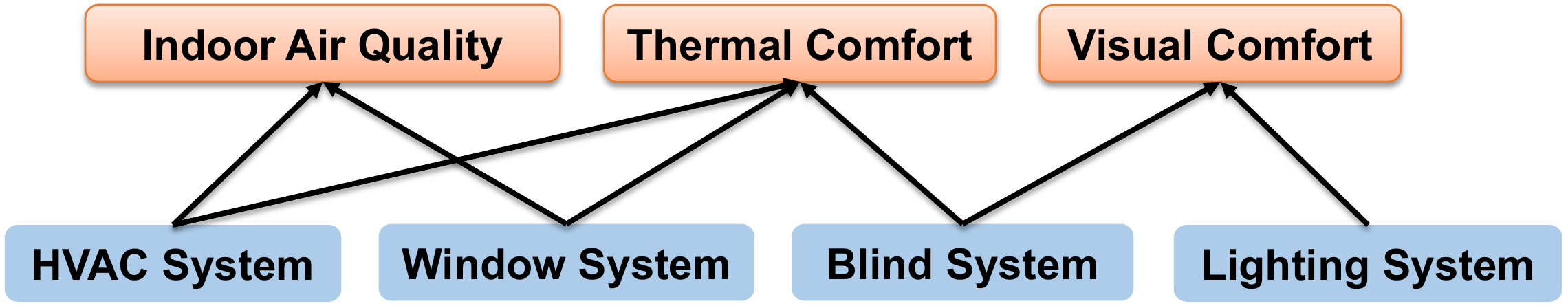}
  \end{center}
  \caption{Relationship between Four Subsystems and Three Human Comfort Metrics}
  \label{system_relationship}
\end{figure}

To achieve more efficient energy management in buildings, we propose to study the joint control problem of four subsystems of a building to meet three human comfort metrics as depicted in Figure~\ref{system_relationship}. 
The energy consumption of a building is determined by four subsystems and their interaction. 
It is challenging to control four subsystems jointly, since they may have opposite outcomes on different human comfort metrics.
For example, opening the window can improve indoor air quality and save the energy consumed by the HVAC system for ventilation, but it may also reduce (in winter) or increase (in summer) indoor temperature. 
To handle the temperature variation caused by the open window, the HVAC system may need to spend more energy rather than the energy saved by natural ventilation.

This paper presents a customized DRL-based control system, named \aliasAPP, which controls four subsystems of a building to meet three human comfort requirements with the best energy efficiency. It leverages all the advantages of DRL-based control, including fast adaptation to new buildings, real-time actuation and being able to handle a large state space. However, to control four subsystems jointly in a unified framework, we need to tackle three main challenges: 

\emph{High-Dimension Control Actions.}
With a uniform DRL framework, \aliasAPP needs to decide a control action for four subsystems jointly and periodically, including the heating/cooling air temperature of the HVAC system, the brightness level of electric lights, the blind slat range and the open proportion of the window. 
Each subsystem adds one dimension in the action space. The goal of \aliasAPP is to select the best action combination $A_{s}$ from the set of all possible combinations $A_{all}$ that meet the requirement of human comfort with the lowest energy consumption. 
Since each subsystem can set its actuator to a large number of discrete values, e.g., we have 66 possible values to set the zone temperature by the HVAC system, the set of all possible action combinations $A_{all}$ is extremely large, i.e., 2,371,842 actions in our case.

To solve this problem, we leverage a novel neural architecture featuring a shared representation followed by four network branches, one for each action dimension.
In addition, from the shared representation, a state value is obtained that links the joint interrelations in the action space, and it is added to the output of the four previous branches. This approach achieves a linear increase in the number of network outputs by allowing independence for each action dimension. 

\emph{Reward Function}.
To explore the potential energy saving energy across four subsystems while considering three human comforts, we formulate this problem into an optimization problem.
We define a reward function in our DRL framework to solve the optimization problem. The novel reward function jointly combines energy consumption, thermal comfort, visual comfort, and indoor air quality, offering better control and more flexibility to meet the unique requirement of users.

\emph{Data Training Requirements}.
While model-free approaches in general, and RL techniques in particular, are very powerful, their main weakness is the amount of data required to train them properly.
The amount of training data should be in proportion to the action space, which in our case it is very large.
This issue is very important since we cannot expect building stakeholders to have years of building data readily available so we can use \aliasAPP. 
Instead, we use a calibrated building simulator combined with weather data that is readily available, in order to generate as much training data as we needed.  
We trained our \aliasAPP system with 10-year of weather data of two areas; one is Merced, CA, and the other one in Chicago, IL, due to their distinct weather characteristics.
The critical point is that this method allows to train \aliasAPP for any building under any weather profile, as long as there is a repository of weather data for the location, and a few months of building data to perform the calibration of the simulator.

We highlight the main contributions of the paper as follows:

\noindent
\textbullet~~To the best of our knowledge, this is the first work that leverages DRL to balance the tradeoff between energy use and human comfort in a holistic manner.

\noindent
\textbullet~~\aliasAPP adopts a special reward function and a new DRL architecture to tackle the challenges imposed by the combined joint control of four subsystems with a very large action space.

\noindent	
\textbullet~~We tackle the issue of data training requirement by adopting a simulation strategy for data generation, and spending effort in calibrating the simulations to make them as close as possible to the target building.  
This allows our system to generate as much data as needed within a finite amount of time.

\section{Related Work}\label{sec:relate_work}

\textbf{Conventional control of the HVAC system}. 
Model predictive control (MPC) models have been developed for HVAC control. It is a planning-based method that solves
an optimal control problem iteratively over a receding time
horizon. Some of the advantages of MPC are that it takes
into consideration future disturbances and that it can handle
multiple constraints and objectives, e.g., energy and comfort \cite{beltran2014optimal}.

However, it can be argued that the main roadblock preventing
widespread adoption of MPC is its reliance on a model \cite{privara2013building, kumar2021building}. By some estimates, modeling can account for up to 75\% of the time and resources required for implementing MPC in practice \cite{rockett2017model}. Because buildings are highly heterogeneous, a custom model is required for each thermal zone or building under control. 

There are two paradigms for modeling building dynamics: physics-based and statistics-based \cite{privara2013building}. Physics-based models, e.g., EnergyPlus, utilize physical knowledge and material properties of a building to create detailed representation of the building dynamics.
A major shortcoming is that such
models are not control-oriented.
Nonetheless, it is not impossible to use such models for
control \cite{atam2016control}. For instance, exhaustive search
optimization is used to derive control policy for an EnergyPlus model \cite{zhao2013energyplus}. Furthermore, physics-based model requires significant modeling
effort, because they have a large number of free parameters to be specified by engineers (e.g., 2,500 parameters for a medium-sized building \cite{karaguzel2011development}); and information required for determining these parameters are scattered in different design documents \cite{gu2014generating}.

Statistical models assume a parametric model form, which may or may not have physical underpinnings, and identifies model parameters directly from data. Dinh et al. \cite{dinh2021milp} propose a hybrid control that combines MPC and direct imitation
learning to reduce energy cost while maintaining a comfortable indoor temperature. While this approach is potentially scalable, a practical problem is that the experimental conditions required for accurate identification of building systems fall outside of normal building operations \cite{agbi2012parameter}.

\textbf{Conventional control of multiple subsystems}. 
Blind system should be considered as an integral part of fenestration system design for commercial and office buildings, in order to balance daylighting requirements versus the need to reduce solar gains. The impact of glazing area, shading device properties and shading control on building cooling and lighting demand was calculated using a coupled lighting and thermal simulation module \cite{tzempelikos2007impact}. The interactions between cooling and lighting energy use in perimeter spaces were evaluated as a function of window-to-wall ratio and shading parameters. 

The impacts of window operation on building performance was investigated \cite{wang2015window} for different types of ventilation systems including natural ventilation, mixed-mode ventilation, and conventional VAV systems in a medium-size reference office building. While the results highlighted the impacts of window operation on energy use and comfort and identified HVAC energy savings with mixed-mode ventilation during summer for various climates, the control for window opening fraction was estimated by experience and is not salable for different kinds of buildings. 

Kolokotsa et al. \cite{kolokotsa2002genetic} develop an energy efficient fuzzy controller based on a genetic algorithm to control four subsystems (HVAC, lighting, window, and blind) and meet the occupant requirements of human comfort. However, the genetic algorithm requires a few minutes to hours to generate one control action and thus is not practical to be used in real building control. 
 
\textbf{RL-based control of the HVAC system}. With the development of deep
learning \cite{liu2020continuous, zhu2021network} and deep reinforcement learning \cite{ding2022drlic, shen2019deepapp}, many works apply RL for HVAC control. RL control can be a “model-free” control method, i.e., an RL agent has no prior knowledge about the controlled process. RL learns an optimal control strategy by “trial-and-error”. Therefore, it can be an online learning method that learns an optimal control strategy during actual building operations. Pedro et al. \cite{fazenda2014using} investigated the application of a reinforcement-learning-based supervisory control approach, which actively learns how to appropriately schedule thermostat
temperature setpoints. However, in HVAC control, online learning may introduce unstable and poor control actions at the initial stage of the learning. In addition, it may take a long time (e.g. over 50 days reported in \cite{fazenda2014using}) for an RL agent to converge to a stable control policy for some cases. Therefore, some studies choose to use an HVAC simulator to the train the RL agent offline \cite{zhang2019whole}. 

Unlike MPC, simulators with arbitrary high complexity can be directly used to train RL agents because of its “model-free” nature. Li et al. \cite{li2015multi} adopt Q learning for HVAC control. Dalamagkidis et al. \cite{dalamagkidis2007reinforcement} design a Linear Reinforcement Learning Controller (LRLC) using linear function approximation of the state-action value function to meet the thermal comfort with minimal energy consumption. However, the tabular Q learning approaches are not suitable for problems with a large state space, like the state of four subsystems. Le et al. \cite{van2019control, le2021deep} propose a control method of air free-cooled data centers in tropics via DRL. Vazquez-Canteli et al. \cite{vazquez2020marlisa} develop a multi-agent RL implementation for load shaping of grid-interactive connected buildings. Ding et al. \cite{ding2020mb2c} design a model-based RL method for multi-zone building control. Zhang et al. \cite{zhang2018practical, zhang2018deep} implement and deploy a DRL-based control method for radiant heating systems in a real-life office building. Gao et al. \cite{gao2020deepcomfort} propose a deep deterministic policy gradients (DDPGs)-based approach for learning the thermal comfort control policy. Although the above works can improve the performance of HVAC control, they only focused on HVAC subsystem.

 \begin{figure*}[!htbp]
	\begin{minipage}[t]{0.32\linewidth}
		\centering
		 \includegraphics[width=2.15in,height=1.35in,angle=0]{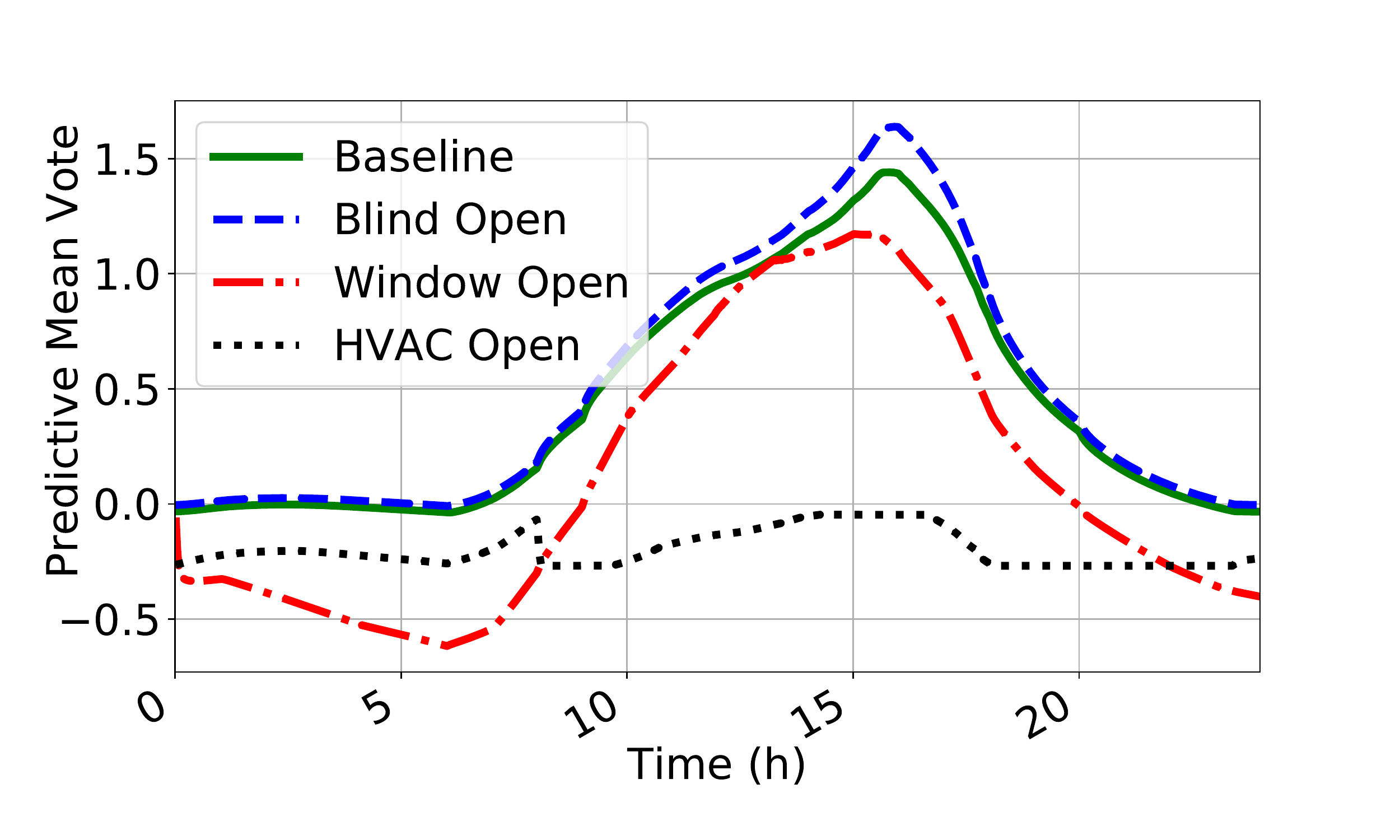}
		\caption{Thermal Comfort, PMV}
		\label{Fig_sub_1}
	\end{minipage}%
		\hspace{1ex}
		\begin{minipage}[t]{0.32\linewidth}
		\centering
		 \includegraphics[width=2.15in,height=1.35in,angle=0]{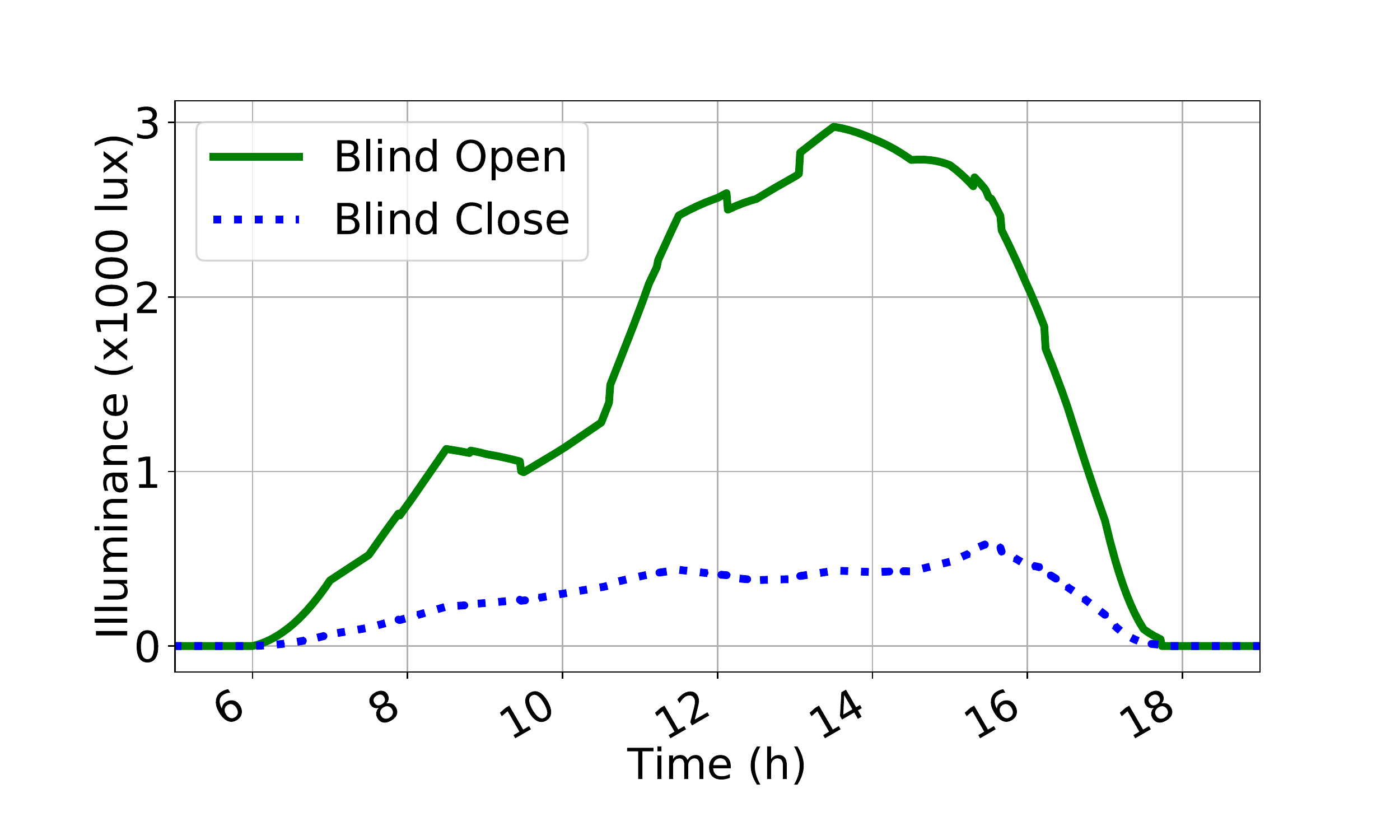}
		\caption{Visual Comfort, Illuminance}
		\label{Fig_sub_2}
	\end{minipage}
	\hspace{1ex}
	\begin{minipage}[t]{0.32\linewidth}
		\centering
		 \includegraphics[width=2.2in,height=1.35in,angle=0]{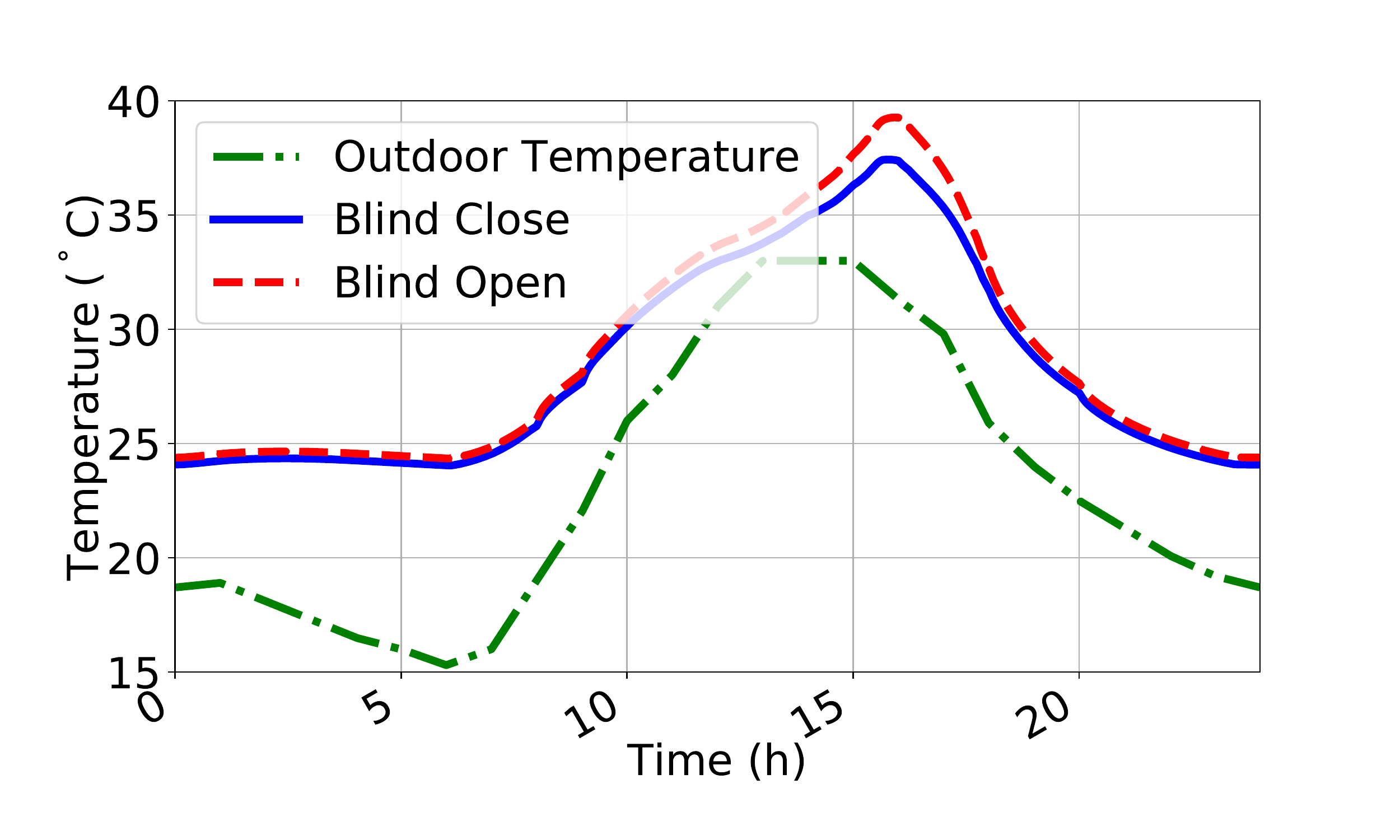}
		\caption{Temperature Effect}
		\label{Fig_sub_3}
	\end{minipage}
\end{figure*}

\section{Motivation}
In this section, we perform a set of preliminary simulations in EnergyPlus~\cite{energyplus} in order to understand the relationships between the different subsystems and their impact on human comfort in a building as described in Figure~\ref{system_relationship}.  This is also used to gain trust that the simulator is being run correctly, with intuitive results that can be understood.

Our goal is to study the effect of different subsystems to three human comfort metrics. A single-floor office building of 100 $m^2$ at Merced, California is modeled. The building is equipped with a north-facing single-panel window of~2~$m^2$ and an interior blind. 
The simulations are conducted with weather data for the month of October.  This is a shoulder season, with outdoor temperatures being a bit cold, but mostly sunny days, i.e. high solar gain.

Figure~\ref{Fig_sub_1} shows the effect of three subsystems on thermal comfort.  
Predictive Mean Vote (PMV) is used to evaluate thermal comfort. 
A PMV value that is close to zero represents the best thermal comfort, with higher positive values meaning people are hot, and lower negative values meaning people are cold. A detailed description of PMV values and ranges will be provided in Section~\ref{subsubHumanComfort}.
The baseline case (green-solid) in Figure~\ref{Fig_sub_1} shows the case when all three subsystems are closed. This case acts like a ``fishtank'' model, where the only effect in the room is due to the solar gain during the day, with no other interactions through any system but the window.  

When only the blind is open (blue-dashed), the PMV value can be affected from 1.45 to 1.75, showing an increase in the temperature due to the increase of solar gain.  This is more prominent in the middle of the day, when the sun is at its apex.   
When the window is open (red-dashed-dot), the PMV value is lowered due to the temperature effect, colder outside air enters the room, producing a colder, more comfortable temperature.  
The HVAC system (black-dot) can maintain the PMV value to an acceptable range (between -0.5 and +0.5) by forcing air to be at the correct temperature through the room vents. 
From the results of Figure~\ref{Fig_sub_1}, we can conclude that all these three subsystems have an obvious impact on thermal comfort. 
Figure~\ref{Fig_sub_2} shows the illuminance measured at a place close to the window from 5 am to 7 pm when the blind is open (green-solid) and the room has natural light.  Illuminance values from 500-1000 lux or higher are acceptable in most environments. We clearly see that with the blind open, the values are within this range for most of the day.

Figure~\ref{Fig_sub_3} shows the indoor temperature when the blind is open (red-dashed) or closed (blue-solid). The outdoor temperature (green-dash-dot) is lower than the indoor temperature, due to the ''fish tank'' effect and the lack of window open or an HVAC system on during the day.
Combining the results from Figures~\ref{Fig_sub_2} and~\ref{Fig_sub_3} we see that the blind system can save the energy consumed by the lighting system by reducing the need of artificial light, but it may also increase the energy used by the HVAC system in order to maintain the load. However, for lower outdoor temperatures in winter, the sunlight through the blind can increase the indoor temperature and save the energy of the HVAC system. 

The simulations are conducted to show some examples of the non-trivial interactions between subsystems and human comfort. It is challenging to quantify the complex relationships among different subsystems and the three human comfort metrics and serves as motivation for our work.

\section{Design of Octopus}
\label{sec:sensingGraph}

In this section, we describe in detail the design of \aliasAPP, including a system overview, DRL-based building control, branching dueling Q-Network, and reward function calculation.

\subsection{\aliasAPP Overview}
The design goal of \aliasAPP is to meet the requirement of human comfort by energy efficient control of four subsystems in a building. 

Our goal is to minimize the energy $E$ consumed by all subsystems in the building, including the energy used in heating/cooling coils to heat and cool the air, the electricity used in the water pumps and flow fans in the HVAC system, electricity used by the lights, and the electricity used by the motors to adjust the blinds and windows.

The value of $E$ is constantly being affected by the vector $A_s$, which is an action combination for four subsystems, which belongs to the vector $A_{all}$ that is all the possible action combinations.

In addition to the minimization of energy, we would like to maintain the human comfort metrics within a particular range. This can be expressed as $P_{min}\leq PMV \leq P_{max}, V_{min}\leq V \leq V_{max}$, and $I_{min}\leq I \leq I_{max}$.

$PMV$ is a parameter that measures thermal comfort; $V$ measures visual comfort; and $I$ measures indoor air quality. The consumed energy $E$ and the human comfort metrics ($PMV$, $V$, and $I$) are determined by the current state of all four subsystems, the outdoor weather and the action we are about to take. They can be measured in real buildings or calculated in a building simulator, like EnergyPlus, after the action is executed.  

The achieved human comfort results should fall into an acceptable range to meet the requirements of users.
We use [$P_{min}$, $P_{max}$], [$V_{min}$, $V_{max}$], [$I_{min}$, $I_{max}$] to present the accepted range for thermal comfort, visual comfort and indoor air quality. 
They can be set by individual users according to their preference, or by facility managers based on building standards.
The details on calculation of the above parameters ($E$, $PMV$, $V$ and $I$), the definition of an action ($A_s$)  and the settings of the human comfort ranges (e.g., [$P_{min}$, $P_{max}$]) will be introduced in Section~\ref{user_demand}.


\begin{figure*}[t]
\begin{center}
  \includegraphics[height=2.5in, width=5.0in]{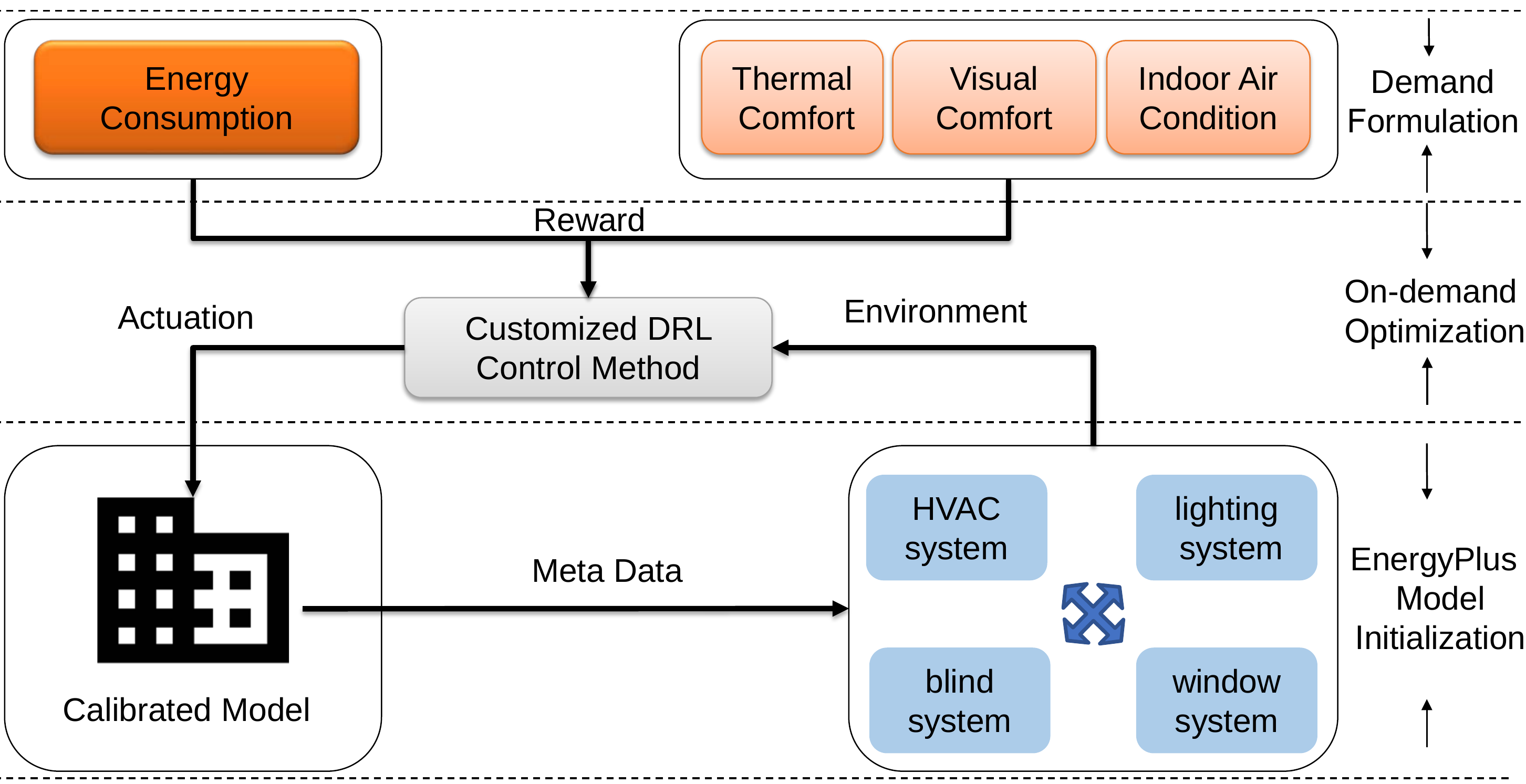}
\end{center}
  \caption{\aliasAPP Architecture with Four Subsystems (including HVAC, lighting, blind and window systems)}
  \label{sensingGraph}
\end{figure*}

Our goal is to find the best $A_s$ from $A_{all}$ for each action interval (15 mins in our implementation). The best $A_s$ should maintain the three human comfort metrics in their acceptable ranges for the entire control interval with the lowest energy consumption ($E$). To achieve this goal, we implement a DRL-based control system for buildings.
Figure~\ref{sensingGraph} shows the overview of \aliasAPP as a building control system. It consists of three layers, i.e., building layer, control layer, and user demand layer. 
The building layer is composed of the real building or a building simulation model, and the sensor data management components.
It provides sensor data to the control layer and executes the control actions generated by the latter. The user demand layer quantifies the user requirement of three human comfort metrics. The range of each human comfort metric is then passed to the control layer, which searches for the optimal control to meet the human comfort ranges with minimal energy consumption.

\subsection{DRL-based Building Control}\label{sec:rl}

\subsubsection{Basics for DRL and DQN}
\begin{figure}[t]
\centering
  \includegraphics[height=1in, width=3in]{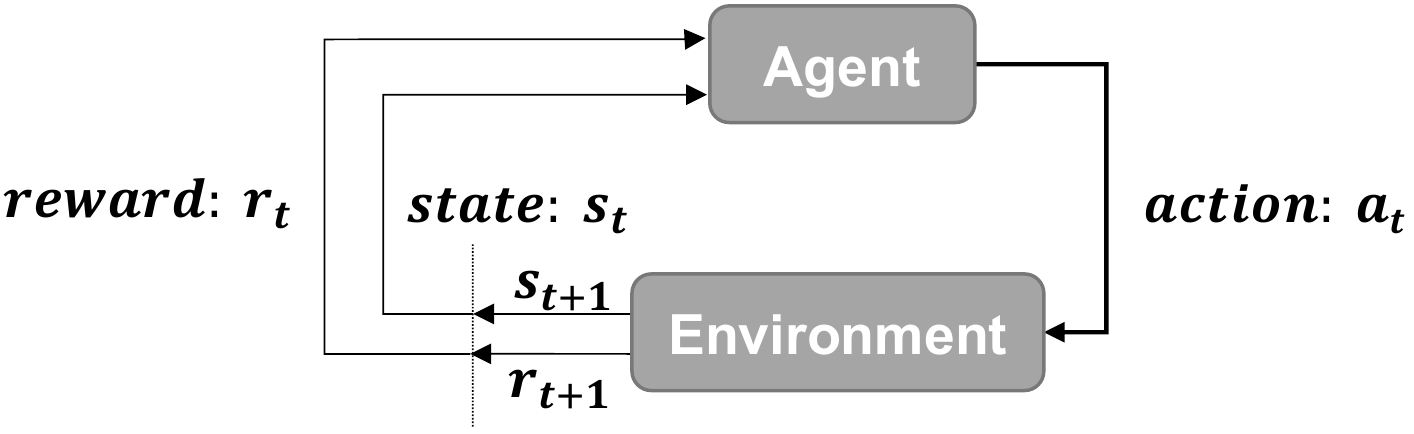}
  \caption{Reinforcement Learning Model.}
  \label{rl}
\end{figure}

In a standard RL framework, as shown in Figure \ref{rl}, an agent learns an optimal control policy by trying different control actions to the environment. 
In our case, the environment is a building simulation model due to the extensive data requirements to train the system.
With DRL, the agent is implemented as a deep neural network (DNN). 
The agent-environment interactions of one step can be expressed as a tuple ($S_{t}$,  $A_{t}$, $S_{t+1}$, $R_{t+1}$), where $S_{t}$ is the environment's state at time $t$, $A_{t}$ is the control action performed by the agent at time $t$, $S_{t+1}$ is the resulting environment' s state after the agent has taken the action, $R_{t+1}$ is the reward received by the agent from the environment. 
The goal of DNN agent training is to learn an optimal control policy to maximize the accumulated returned reward by taking different control actions.

\subsubsection{State in \aliasAPP}
The state is what the DRL agent takes as input for each control step. In this study, the state is a stack of the current and historical observations, as shown below:
\begin{equation}
\begin{array}{l}
S =\left \{ob_{t}, ob_{t-1}, ..., ob_{t-n}  \right \},
\end{array}
\end{equation}
where $t$ is the current time step, $n$ is the number of the historical time steps to be considered, and each $ob$ consists of the following 15 items: outdoor air temperature ($^{\circ}$C), 
outdoor air relative humidity (\%), indoor air temperature($^{\circ}$C), indoor air relative humidity (\%), diffuse solar radiation ($W/m^{2}$), direct solar radiation ($W/m^{2}$), solar incident angle ($^{\circ}$), wind speed (m/s), wind direction (degree from north), 
average PMV (\%), heating setpoint of the HVAC system ($^{\circ}$C), cooling setpoint of the HVAC system ($^{\circ}$C), the dimming level of lights (\%), the window open percentage (\%), and the blind open angle ($^{\circ}$). All the values we can be calculated by the EnergyPlus simulation model. Min-max normalization is used to convert each item to a value within 0-1.

\begin{figure*}[t]
\centering
\includegraphics[width=6.in, height=1.6in]{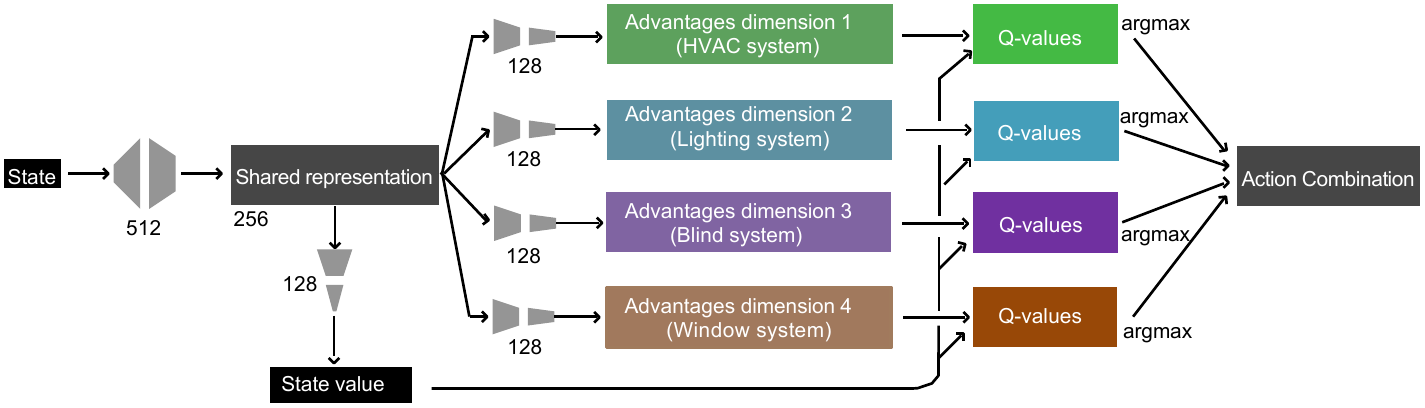}
\caption{The Specific Action Branching Network Implemented for the Proposed BDQ Agent}
\label{branching_network}

\end{figure*}

\subsubsection{Action in \aliasAPP}
\label{action_section}
The action is how the DRL agent controls the environment. Given the state, we want the agent to find the most suitable action combinations among HVAC, lighting, blind and window system to balance energy consumption and three human comfort metrics. There are four action dimensions when considering these four subsystems, represented as 
\begin{equation}
\begin{array}{l}
A_{t} =\left \{H_{t}, L_{t}, B_{t}, W_{t} \right \},
\end{array}
\label{action_combination}
\end{equation}
where $A_{t}$ is the action combination of four subsystems at time $t$. $H_{t}$ is the temperature set-point of the HVAC system, which can be set to 66 values. $L_{t}$ is the dimming level of electric lights. $B_{t}$ is the blind slat angle. The range of blind slat can be adjusted from 0 $^{\circ}$  $\sim$ 180 $^{\circ}$. $W_{t}$ is the open percentage of the window. Each of the above three actuation parameters can be set to 33 values in our current implementation to achieve a proper balance between control granularity and calculation complexity. According to Equation \ref{action_combination}, the total number of possible actions in the action space is 2,371,842 ($66\times33\times33\times33$). 
Existing DRL architectures, like Deep Q-Network (DQN) in~\cite{wei2017deep} and Asynchronous Advantage Actor-Critic (A3C) in~\cite{zhang2018practical}, cannot work efficiently in our problem, because the large number of actions requires to be explicitly represented in the agent DNN network and it will significantly increase the number of DNN parameters to be learned and consequently the training time~\cite{van2016deep}. To solve this problem, we leverage a novel neural architecture featuring a shared representation followed by four network branches, one for each action dimension.

\subsubsection{Reward Function in \aliasAPP}
\label{reward_section}
Reward illustrates the immediate evaluation of the control effects for each action under a certain state. Both human comfort and energy consumption should be incorporated. To define the reward function, a common approach is to use the Lagrangian Multiplier function~\cite{ito2008lagrange} to first convert the constrained formulation into an unconstrained one:
\begin{equation}
  \begin{aligned}
    R = - [ \rho _{1}Norm(E)+ \rho _{2}Norm(T_{c}) \\
        + \rho _{3}Norm(V_{c}) + \rho _{4}Norm(I_{c}) ],
  \end{aligned}
  \label{lagrangian}
\end{equation}
where $\rho _{1}$, $\rho _{2}$, $\rho _{3}$ and $\rho _{4}$ are the Lagrangian multipliers. $E$ is energy consumption, $T{c}$ is thermal comfort, $V{c}$ is visual comfort and $I{c}$ is Indoor air quality. $Norm(x)$ is a normalization process, i.e., $Norm(x)$ = (x - $x_{min}$ )/($x_{max}$ - $x_{min}$) to transform energy and three human comfort to the same scale. This reward function merges the objective (e.g. energy consumption) and constraint satisfaction (e.g. human comfort). The reward consists of four parts, namely, the penalty for the energy consumption of the HVAC and lighting system, the penalty for the occupants' thermal discomfort, the penalty for the occupants' visual discomfort and the penalty for the occupants' indoor air condition discomfort. Specifically, the reward should be less, if more energy is consumed by the HVAC system or the occupants feel uncomfortable about the building thermal, visual and indoor air condition. The details about how to define and formulate energy consumption $E$, thermal comfort $T{c}$, visual comfort $V{c}$ and indoor air condition $I{c}$ are explained in Section~\ref{user_demand}.

\subsection{Branching Dueling Q-Network}
\label{bdq_network}

To solve the high-dimensional action problem described in Section~\ref{action_section}, \aliasAPP adopts a Branching Dueling Q-Network (BDQ), which is a branching variant of the dueling Double Deep Q-Network (DDQN). BDQ is a new neural architecture featuring a shared decision module followed by several network branches, one for each action dimension. BDQ can scale robustly to environments with high dimensional action spaces and even outperform the Deep Deterministic Policy Gradient
(DDPG) algorithm in the most challenging task \cite{tavakoli2018action}. In our current implementation, we use a simulated building model developed in EnergyPlus as the environment for training and validation. 
Our BDQ-based agent interacts with the EnergyPlus model. At each control step, it processes the state (building and weather parameters) and generates a combined action set for four subsystems. 

Figure~\ref{branching_network} demonstrates the action branching network of BDQ agent. When a state is inputted, the shared decision module computes a latent representation that is then used for the calculation of the state value and the output of the network (Advantages dimension in Figure~\ref{branching_network}) for each dimension branch. The state value and the factorized advantages are then combined, via a special aggregation layer, to output the Q-values for each action dimension. These Q-values are then queried for the generation of a joint-action tuple. The weights of the fully connected neural layers are denoted by the gray trapezoids and the size of each layer (i.e. number of units) is depicted in the figure.

\begin{algorithm}[t]
    \caption{The Training Process of Our BDQ-Based Agent}
    \label{training-algo}
    \KwIn{The range of human comfort metrics and maximum acceptable energy consumption}
    \KwOut{A trained DRL agent}    
    Initialize BDQ's prediction Q with random weights $\theta$\;
    Initialize BDQ's target $Q^{-}$ with weight $\theta^{-} = \theta$ \;
    \For{episode =0,1,...,M}
    {
      Obtain the initial state $S_{t}$ and $A_{t}$ randomly\;
      \For{control time step t = 0,1,...,T}
      {
        Update $ H_{t}, L_{t}, B_{t}, W_{t}$ by the control action, $A_{t}$\;
        Calculate reward $R_{t+1}$\ by Equation \ref{lagrangian}\;
        Obtain current state observation $S_{t+1}$\;
        Store ($S_{t}$, $A_{t}$, $S_{t+1}$, $R_{t+1}$) in reply memory $\Lambda$\;
        Draw mini-batch sample transitions from $\Lambda$\;
        Calculate the target vector and update weights in neural network Q \;
        Update target network $Q^{-}_d(s,a_{d})$ using Equation \ref{update_q_value}\;
        Perform greedy descent iteratively to tune BDQ by Equation \ref{loss_function}.\
      }
    }
\end{algorithm}

\textbf{Training Process:}
The training process of the BDQ-based control agent is outlined in Algorithm~\ref{training-algo}. At the beginning, we first initialize a neural network $Q$ with random weight $\theta$. 
Another neural network $Q^{-}$ with the same architecture is also created.
The outer "for" loop controls the number of training episodes, and the inner "for" loop performs control at each control time step within one training episode. During the training process, the recent transition tuples ($S_{t}$, $A_{t}$, $S_{t+1}$, $R_{t+1}$) are stored in the replay memory $\Lambda$\, from which a mini-batch of samples will be generated for neural network training.
The variable \emph{$A_{t}$} stores the control action in the last step, and $S_{t}$ and $S_{t+1}$ represent the building state in the previous and current control time steps, respectively. At the beginning of each time slot t, we first update four actions and obtain the current state $S_{t+1}$. In line 7, the immediate reward $R_{t+1}$ is calculated by Equation~\ref{lagrangian}. A training mini-batch can be built by randomly drawing some transition tuples from the memory. 

We calculate the target vector and update the weights of the neural network $Q$ by using an Adam optimizer for every control step $t$. Formally, for an action dimension $d\in{1,...N}$ with $n$ discrete actions, a branch's Q-value at state $s\in S$ and with action $a_{d} \in A_{d}$ is expressed in terms of the common state value $V(s)$ (the result of the shared representation layer in Figure~\ref{branching_network}) and the corresponding (state-dependent) action advantage $A_{d}(s,a_{d})$ of each branch (the result of the each advantage dimension in Figure~\ref{branching_network}) by:
\begin{equation}
\begin{array}{l}
Q_d(s,a_{d}) = V(s) + (A_{d}(s,a_{d})- \frac{1}{n} \sum_{a_{d}^{'}\in A_{d}}A_{d}(s,a_{d}^{'})).
\end{array}
\label{q_value}
\end{equation}

The target network $Q^{-}$ will be updated with the latest weights of the network $Q$ every $c$ control time steps. $c$ is set to 50 in our current implementation. $Q^{-}$ is used for inferring the target value for the next $c$ control steps.
We use $y_{d}$ to represent the maximum accumulative reward we can obtain in the next $c$ steps. $y_{d}$ can be calculated by temporal-difference (TD) targets in a recursive fashion:
\begin{equation}
\begin{array}{l}
y_{d} = R + \gamma \frac{1}{N}\sum_{d}Q_{d}^{-}(s^{'}, \arg\max\limits_{a_{d}^{'}\subseteq A_{d}} Q_{d}(s^{'},a_{d}^{'})),
\end{array}
\label{update_q_value}
\end{equation}
where $Q_{d}^{-}$ denoting the branch $d$ of the target network $Q^{-}$; $R$ is the reward function result; and $\gamma$ is discount factor.

Finally, at the end of the inner "for" loop, we calculate the following loss function every $c$ control steps:
\begin{equation}
\begin{array}{l}
L = \mathbb{E}_{(s,a,r,s_{'})} \sim D\left [\sum_{d} (y_{d}-Q_{d}(s,a_{d}))^{2} \right ],
\end{array}
\label{loss_function}
\end{equation}
where $D$ denotes a (prioritized) experience replay buffer and \emph{a} denotes the joint-action tuple ($a_{1}, a_{2}, ..., a_{N}$). The loss function $L$ should decrease as more training episodes are performed.

\subsection{Reward Calculation}
\label{user_demand}
This section describes how we calculate the reward function in Equation \ref{lagrangian}, including energy cost $E$, thermal comfort $T$, visual comfort $V$ and indoor air condition $I$.

\subsubsection{Energy Consumption} \label{Energy}
The energy consumption of a building includes heating coil power $P_{h}$ and cooling coil power $P_{c}$ and fan power $P_{f}$ from the HVAC system and electric light power $P_{l}$ from the lighting system. We calculate the reward function for energy consumption $E$ during a time slot as
\begin{equation}
\begin{aligned}
E = (P_{h} + P_{c} + P_{f} + P_{l})
\end{aligned}
\end{equation}

The heating and cooling coil are used to cool or heat the air and the fan is used to distribute the heating air or cooling air to the zone. The electric lights are used for normal work in the zone. They are calculated by EnergyPlus simulator in our training and evaluation. In our current implementation, we ignore the power consumed by the water pumps and the motors to adjust blinds and windows, because it is relatively small compared with the power consumption of the HAVC system or the lighting systems, and can be safely ignored (less than 1\% total). 

\subsubsection{Human Comfort}
\label{subsubHumanComfort}
We define and explain the measurement of the three human comfort metrics.

\textbf{Thermal Comfort:} It is determined by the index PMV (Predictive Mean Vote) that is calculated by Fanger's equation~\cite{fanger1970thermal}. 
PMV predicts the mean thermal sensation vote on a standard scale for a large group of persons. The American Society of Heating Refrigerating and Air Conditioning Engineers (ASHRAE) developed the thermal comfort index by using coding -3 for cold, -2 for cool, -1 for slightly cool, 0 for natural, +1 for slightly warm, +2 for warm, and +3 for hot. PMV has been adopted by the ISO 7730 standard~\cite{fanger1984moderate}. The ISO recommends maintaining PMV at level 0 with a tolerance of 0.5 as the best thermal comfort. We calculate the reward function for thermal comfort $T_{c}$ during a time slot as
\begin{equation}
T_{c} =
\left\{
             \begin{array}{lr}
              0,         &PMV \leq P \\
             |PMV-P|,    &PMV > |P| \\

             \end{array}
\right.
\end{equation}

The occupants can feel comfort when PMV value is within an acceptable range. We denote the range as $[-P, P]$, where P is the threshold for PMV value. If the PMV value lies
within $[-P, P]$, it will not incur a penalty. Otherwise, it will incur a penalty for the occupants' dissatisfaction with the building thermal condition. There are six primary factors that directly affect thermal comfort that can be grouped in two categories: personal factors - because they are characteristics of the occupants - and environmental factors - which are conditions of the thermal environment. The former are metabolic rate and clothing level, the latter are air temperature, mean radiant temperature, air speed and humidity. The PMV personal factors parameters are shown in Table \ref{PMV_Constants}. The PMV personal factors environmental factors are obtained in real time from EnergyPlus.


\begin{table}[t]
  \renewcommand\arraystretch{1.2}
  \small
  \caption{PMV Constants}
  \centering
  \begin{tabular}{c|c|c}
\hline
Parameter& Value & Units\\
\hline
Metabolic rate & 70& $W/m^{2}$\\
\hline
Clothing Level & 0.5 & clo\\
\hline
  \end{tabular}
  \label{PMV_Constants}
\end{table}

\textbf{Visual Comfort:} The research on visual comfort is dominated by studies analyzing the presence of an adequate amount of light where discomfort can be caused by either too low or too high level of light as glare. In this paper, the major glare metric is illuminance range~\cite{pritchard2014lighting}. The illuminance source includes daylight and electrical light. Thus, the main subsystems that can have an impact on visual comfort are blind system and lighting system. We calculate the reward function for visual comfort $V_{c}$ during a time slot as
\vspace{-0.05in}
\begin{equation}
V_{c} = 
\left\{
             \begin{array}{lr}
             -F - M_{L},  &F< M_{L}  \\
             0,           &M_{L} \leq F  \leq M_{H}  \\
              F - M_{H},  &F> M_{H}\\
             \end{array}
\right.
\end{equation}

The occupants can feel comfort when illuminance value F is within an acceptable range. We denote the range as [$M_{L}$, $M_{H}$], where M is the threshold for illuminance value. If the illuminance value lies within [$M_{L}$, $M_{H}$], it will not incur a penalty. Otherwise, it will incur the penalty for the occupants' dissatisfaction with the building illuminance condition.

\textbf{Indoor Air Quality:} Carbon dioxide (CO$_{2}$) concentration in a building is used as a proxy for air quality~\cite{emmerich2001state}. The carbon dioxide concentration comes from building's users. There are various other sources of pollution (NOx, Total Volatile Organic Compounds (TVOC), respirable particles, etc.). Ventilation is an important means for controlling indoor air quality (IAQ) in buildings~\cite{ashrae_indoor_air_condition}. Ventilation in this work mainly comes from the HVAC system and the window system. We calculate the reward function for indoor air condition $I_{c}$ during a time slot as
\vspace{-0.05in}
\begin{equation}
I_{c} = 
\left\{
             \begin{array}{lr}
             -C - A_{L} , & C < A_{L}  \\
             0,           & A_{L}  \leq C  \leq A_{H}  \\
             C - A_{H},   & C > A_{H}\\
             \end{array}
\right.
\end{equation}

The occupants can feel comfort when carbon dioxide concentration value C is within an acceptable range. We denote the range as [$A_{L}$, $A_{H}$], where A is the threshold for dioxide concentration value. If the dioxide concentration value lies within [$A_{L}$, $A_{H}$], it will not incur a penalty. Otherwise, it will incur a penalty for the occupants' dissatisfaction with the building indoor air quality.

\section{Implementation of Octopus}
\label{sec:Implementation}
In this section, we illustrate in detail the implementation of \aliasAPP including platform setup, HVAC modeling and calibration, and \aliasAPP training.

\subsection{Platform Setup}

\begin{figure}[t]
\begin{center}
\includegraphics[height=2.0in, width=2.6in]{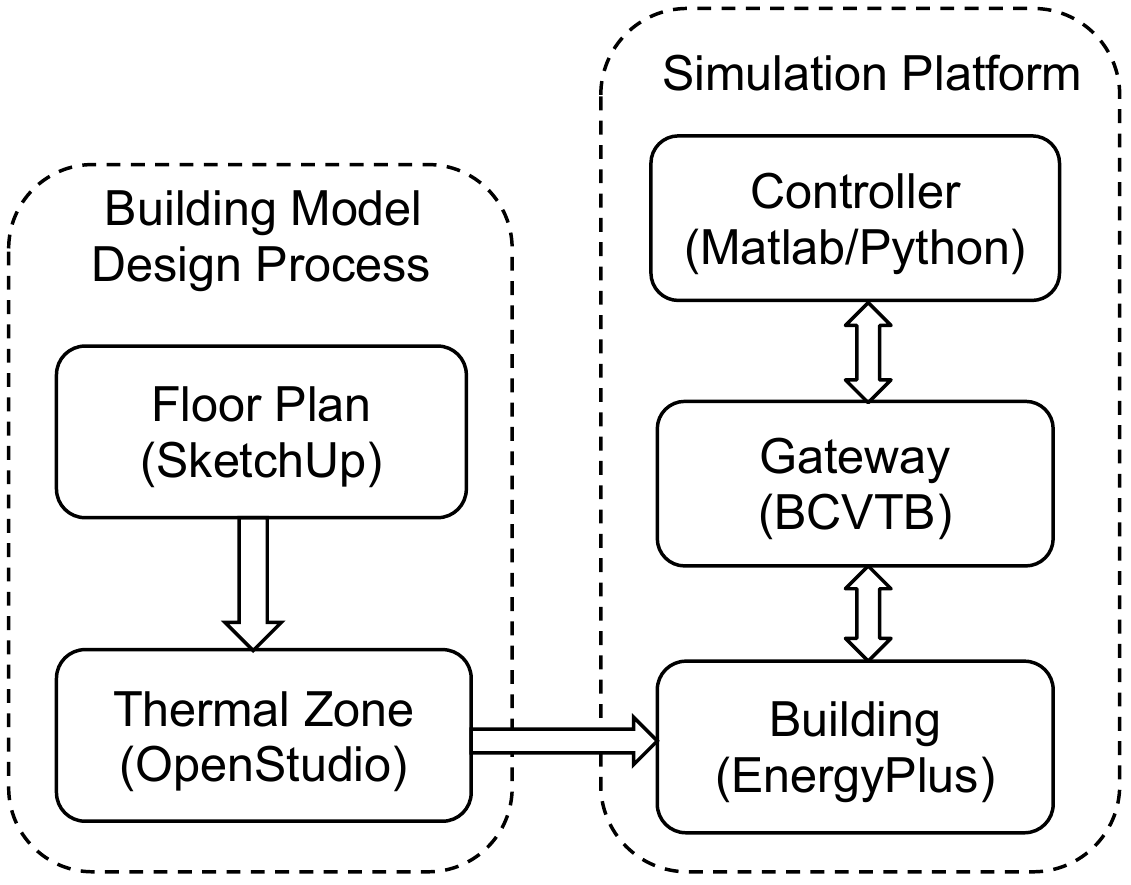}
\end{center}
\caption{Workflow of Octopus}
\label{platform}
\end{figure}

Figure. \ref{platform} shows a conceptual flow diagram of our building simulation and control platform. 
Our building model is rendered using SketchUp~\cite{sketchup}.  It replicates a LEED Gold Certified Building in our University Campus. Using OpenStudio, the HVAC, lighting, blind and window system are installed in the building/zones. The control scheme - \aliasAPP is implemented using Tensorflow, which is an open-source machine learning library for Python. Using the Building Control Virtual Test Bed (BCVTB), a Ptolemy II platform that enables co-simulation across different models~\cite{wetter2011co}, we implement the control of each zone temperature set points, blinds, lighting and window schedule during each action time in EnergyPlus for our Building alongside weather data. \aliasAPP is modeled using EnergyPlus version 8.6~\cite{energyplus}.  We train \aliasAPP based on 10-year weather data from two different cities, Merced, CA and Chicago, IL due to their distinct weather characteristics. The weather data for Merced has intensive solar radiation and large variance in temperature, while Chicago is classified as hot-summer humid continental with four distinct seasons.  To train our model, we define an ``episode'' as one inner for loop of Algorithm~\ref{training-algo}.

\subsection{Rule Based Method}\label{rule_based_method}
We implement a rule-based method based on our current campus building control policy.  This policy was first set up at commissioning time by a mechanical engineering company, and then it was further optimized by two experienced HVAC engineers when going over the LEED certification process. 

First, we assign different zone temperature setpoints. Each zone has a separate heating and cooling setpoint. The heating setpoint is set to 70 $^{\circ}$F, and the cooling setpoint to 74 $^{\circ}$F during the warm-up stage. The cooling setpoint is limited between 72$^{\circ}$F and 80$^{\circ}$F, and the heating setpoint is limited between 65$^{\circ}$F and 72$^{\circ}$F. Second, we set control restrictions and actuator limits and control inputs are subject to the following constraints: the heating setpoint should not exceed the cooling setpoint minus 1 $^{\circ}$F. The adjustment will move both the existing heating and cooling setpoints upwards or downwards by the same amount unless the limit has been reached. Third, for the control Loops: two separate control loops operate to maintain space temperature at setpoint, the Cooling Loop and the Heating Loop. Both loops are continuously active.

\subsection{HVAC System Description}
\begin{figure}[t]
\begin{center}
  \includegraphics[height=1.7in, width=3.0in]{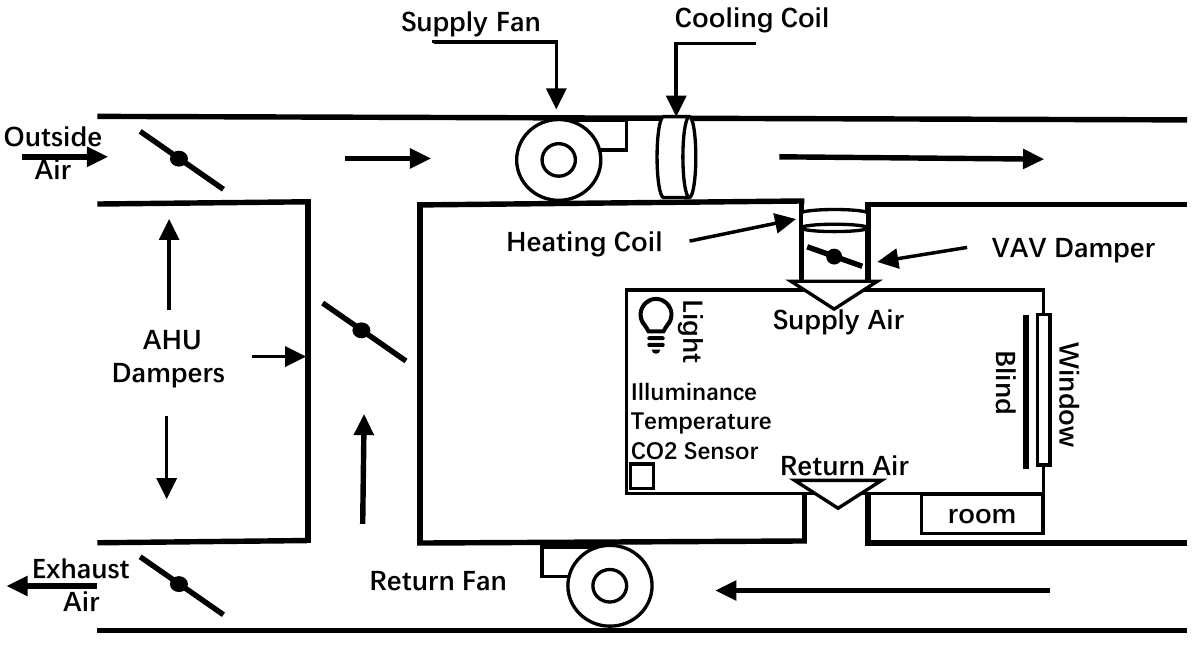}
  \end{center}
  \caption{HVAC Single Duct VAV Terminal Reheat Layout.}
  \label{hvac_single}
\end{figure}

The HVAC system we modeled is a single duct central cooling HVAC with
terminal reheat as shown in Figure \ref{hvac_single}. The process begins at the supply fan in
the air handler unit (AHU), which supplies air for the zone. The supply
fan’s air first goes through a cooling coil, which cools the air to the minimum
required temperature needed for the zone. Before air enters a zone, the
air passes through a variable air volume (VAV) unit that regulates the amount
of air that flows into a zone. Terminal reheat occurs when the heating coil
increases the temperature before discharging air into a zone. A discharge setpoint
temperature is selected for each zone and the VAV ensures that the air is heated
to this temperature for each zone. The air supplied to the zone is mixed with the
current zone air, and some of the air is exhausted out of the zone to maintain a
constant static pressure. The return air from each zone is mixed in the return
duct, and then portions of it may enter the economizer.

\subsection{HVAC Modeling and Calibration}
\label{sec:Calibration}

\begin{figure}[t]
\begin{center}
  \includegraphics[height=2.5in, width=3.0in]{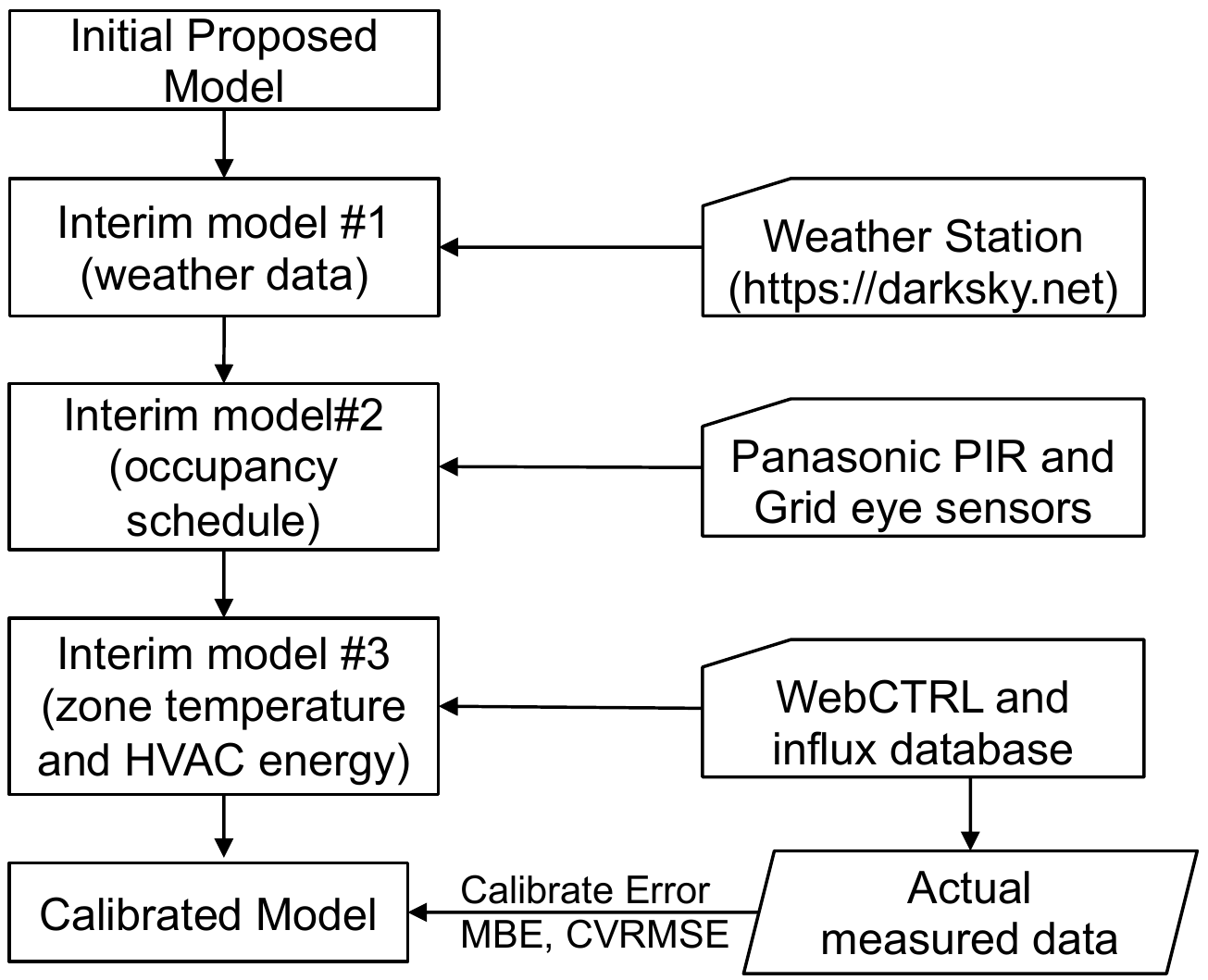}
  \end{center}
  \caption{Building Model Calibration Process.}
  \label{calibration_process}
\end{figure}

The purpose of the calibration is to ensure the building energy model can generate energy use results close to the measured values in the target building using actual inputs, including weather, occupancy schedule, and the HVAC system parameters and controls. 

The building model calibration process is shown in Figure \ref{calibration_process}. The first step of the calibration is to collect the real weather data from a public weather station for the period to be tested. We use a Dark Sky's API, a public weather website, to collect real weather data for three months. The second step is to replace the default occupancy schedules in the simulator with the actual occupancy schedules collected from the real target building using ThermoSense~\cite{beltran2013thermosense}.  This system was installed in the target building on our campus and allows the collection of fine grain occupancy data at the zone level in the building, allowing the evaluation using accurate occupancy patterns.  We used the hourly occupancy data from 3 months as the occupancy schedule in our simulated building by EnergyPlus. The third step is to calibrate certain system and control parameters to match those in the target building we want to replicate.  This involves multiple issues, including (a) the selection of the parameters to be calibrated, (b) the range of those parameters, and (c) the step used within the range.  In our work, we use an N-factorial design with 5 parameters and ranges to be tested based on operational experience.   We tested different combinations of HVAC system parameters (Infiltration rate) and control (mass flow rate, heating, and cooling setpoints) and found the combination that minimized the calibrated error (see below). The selected calibration parameters are listed in the Table~\ref{calibration_parameter} with their calibration ranges and value selected.  The final step is to compare the calibrated error between the calibrated model and the actual measured zone temperature and energy consumption stored in the operational building database. The whole calibration process of modeling our building takes nearly one month.


\begin{table}[t]
  \renewcommand\arraystretch{1.2}
  \small
  \caption{Model Calibration Parameters}
  \centering
  \begin{tabular}{c|c|c}
    \hline
     Parameter &   Range& Adoption\\
    \hline
    Infiltration Rate   & 0.01 $m^{3}$ $\sim$ 0.5 $m^{3}$& 0.05 $m^{3}$ \\
    \hline
    Window Type   & Single/Double Pane & Single  \\
    \hline
    Window Area   & $1m^{2}$ $\sim$ $4m^{2}$& $2m^{2}$  \\
    \hline
    Window Thickness   & $3mm$ $\sim$ $6mm$ &  $3mm$  \\ 
    \hline
    Fan Efficiency   &0.5 $\sim$ 0.8 &  0.7 \\
    \hline
    Blind Type & Interior/Exterior Blind &  Interior \\
    \hline
    Blind Thickness & $1mm$ $\sim$ $6mm$ &  $1mm$  \\
    \hline
  \end{tabular}
  \label{calibration_parameter}
\end{table}


\begin{table}[t]
  \renewcommand\arraystretch{1.2}
  \small
  \caption{Modeling Error after Calibration}
  \centering
  \begin{tabular}{c|c|c}
    \hline
    &   MBE &   CVRMSE\\
    \hline
    February (hourly temperature) &-1.48\%   & 5.32\% \\
    \hline
    March  (hourly temperature) &-0.26\%   & 4.95\%  \\
    \hline
    April  (hourly temperature) &1.20\%    & 5.06\%  \\
    \hline
    May  (hourly temperature) &0.48\%  &   4.38\%  \\
    \hline
    February - May(monthly energy) &-3.83\%  &  12.33\%  \\
    \hline
  \end{tabular}
  \label{calibration}
\end{table}

\begin{table*}[t]
  \renewcommand\arraystretch{1.2}
  \small
  \caption{Human Comfort Statistical Results for Rule Based, DDQN-HVAC and \aliasAPP Schemes}
  \centering
  \begin{tabular}{c|c|c|c|c|c|c|c|c|c|cc}
    \cline{1-11}\multirow{2}{*}{Location}
    &\multirow{2}{*}{Method}  & \multirow{2}{*}{Metric} &\multicolumn{2}{c|}{PMV}&  
    \multicolumn{2}{c|}{\shortstack{Illuminance (lux)} }
 
    & \multicolumn{2}{c|}{\shortstack{CO$_{2}$ Concentration\\ (ppm)}}&\multicolumn{2}{c}{\shortstack{Energy Consumption\\ (kWh)}} \\
    \cline{4-11}
    &  & & January&July&January&July&January&July&January &July\\
    \cline{1-11}
    \multirow{9}{*}{\shortstack{Merced}}
    & \multirow{3}{*}{\shortstack{Rule Based\\Method}} & Mean & 0.03&-0.25&576.78&646.45&623.61&668.03&\multirow{3}{*}{1990.99} &\multirow{3}{*}{3583.03}\\
    \cline{3-9}
    &  & Std& 0.11& 0.13  &   152.54& 157.11& 120.64&181.22& & \\
    \cline{3-9}				
    &  & Violation rate & 0& 2\%  &  0.94\%& 0& 	0.3\%&3.629\%& & \\
    \cline{2-11}	

    & \multirow{3}{*}{
    \shortstack{DDQN-HVAC\\ \cite{zhang2018practical}}
    } & Mean &-0.19 &0.28 & 576.78& 646.45& 625.62&648.01 &\multirow{3}{*}{1859.10} &\multirow{3}{*}{3335.58}\\
    \cline{3-9}				
    &  & Std& 0.21&  0.11  &  	152.54  & 157.11& 122.62 &120.57  & &\\
    \cline{3-9}					
    &  & Violation rate &2.99\% &4.4\%  & 0.94\% & 0&0 & 0.2\% & &\\
 
    \cline{2-11}

    & \multirow{3}{*}{\shortstack{\aliasAPP}} & Mean &-0.31 &0.27 &	587.12 &	569.88&594.77	& 612.33&\multirow{3}{*}{1756.24} &\multirow{3}{*}{2941.46} \\
    \cline{3-9}			
    &  & Std&0.2 &	0.10 & 382.27 &75.83 &111.59 &110.35 & & \\
    \cline{3-9}			
    &  & Violation rate&	5.7\%  & 2.5\% &  0.26\%& 0.2\%& 1.31\%&	0.33\% & & \\
    \cline{1-11}

    \multirow{9}{*}{Chicago}			
    & \multirow{3}{*}{\shortstack{Rule Based\\Method}} & Mean &-0.28 &-0.15 & 583.27& 637.07& 610.26&638.33 &\multirow{3}{*}{3848.61} &\multirow{3}{*}{3309.56}\\
    \cline{3-9}				
    &  & Std& 0.11&  0.02  &  	163.96  & 151.37& 63.94 &151.37  & &\\
    \cline{3-9}					
    &  & Violation rate &3.09\% &0  & 1.1\% & 0&0 & 0 & &\\
    \cline{2-11}						
    & \multirow{3}{*}{\shortstack{DDQN-HVAC\\ \cite{zhang2018practical}}} & Mean &-0.32 &0.24 &583.27 & 637.07&612.74 &649.32 &\multirow{3}{*}{3605.21}&\multirow{3}{*}{3078.67}  \\
    \cline{3-9}
    &  & Std&0.08 &	0.07 &163.96 &	151.37 &65.09	 &90.16&& \\
    \cline{3-9}		
    &  & Violation rate & 	3.7\%& 2.9\% &	1.1 \% &	0 & 0 & 	0 && \\
    \cline{2-11}	
    & \multirow{3}{*}{\aliasAPP} & Mean &-0.4 &0.29 &598.34 & 544.09&640.31 &633.71 &\multirow{3}{*}{3496.54}&\multirow{3}{*}{2722.03}  \\
    \cline{3-9}
    &  & Std&0.1 &	0.11 &259.88 &	55.37 &99.85	 &111.04&& \\
    \cline{3-9}		
    &  & Violation rate & 	4.2\%& 1.47\% &	1.6 \% &	0 & 1\% & 	1.31\% && \\
    \cline{1-11}

  \end{tabular}
  \label{human_comfort}
\end{table*}

ASHRAE Guideline 14-2002~\cite{guideline2002guideline} defines the evaluation criteria to calibrate BEM models. According to the Guideline, monthly and hourly data can be used for calibration. Mean Bias Error (MBE) and Coefficient of Variation of the Root Mean Squared Error (CVRMSE) are used as evaluation indices. The guideline states that the model should have an MBE of 5\% and a CVRMSE of 15\% relative to monthly calibration data. If hourly calibration data are used, these requirements should be 10\% and 30\%, respectively. In our case, hourly data is used to calculate the error metrics for the average zone temperature. We choose monthly data to calculate energy error metrics because energy data can only be obtained monthly. The calibration results for zone temperature and energy consumption are shown in Table~\ref{calibration}. It is shown that less than 2\% NMBE and less than 6\% CVRMSE for the zone temperature can be achieved with the optimal parameter setting. We found that both the CVRMSE for the monthly heating and cooling energy demand is relatively large, but the NMBE and CVRMSE are still within the acceptable range. This means the model can achieve accurate calculation for the monthly energy.

\subsection{\aliasAPP Training}


\begin{table}[t]
  \renewcommand\arraystretch{1.2}
  \small
  \caption{Parameter Settings in DRL Algorithms}
  \centering
  \begin{tabular}{c|c|c|c}
    \hline
    $\bigtriangleup t_{c}$ & 15 m &$\beta_{1}$ & 0.9\\
    \hline
    Minibatch Size &  64 & $\beta_{2}$ & 0.999\\
    \hline
    Learning Rate & $10^{-4}$& Action Dimension& 35040 \\
    \hline
    $\gamma$ & 0.99 & Action Space & $2.37* 10^{7}$\\
    \hline
  \end{tabular}
  \label{parameter_drl1}
\end{table}

10-year weather data for training from the two locations tested (Merced, CA and Chicago, IL) is randomly divided, with eight years used for training and the remaining two years used for testing. The parameter settings in our DRL Algorithms are shown in Table \ref{parameter_drl1}. In our implementation of \aliasAPP, we use the Adam optimizer~\cite{kingma2014adam} for gradient-based optimization with a learning rate of $10^{-4}$. We train the agent with a minibatch size of 64 and a discount factor $\gamma$ = 0.99. The target network is updated every $10^{3}$ time steps. We use the rectified non-linearity (or ReLU)~\cite{glorot2011deep} for all hidden layers and linear activation on the output layers. The network has two hidden layers with 512 and 256 units in the shared network module and one hidden layer per branch with 128 units. The weights are initialized using the Xavier initialization~\cite{glorot2010understanding} and the biases were initialized to zero.

We used the prioritized replay with a buffer size of $10^{6}$ and linear
annealing of $\beta$ from $\beta_{0}$ = 0.4 to 1 over $2$ x $10^{6}$
steps. While an $\epsilon-$greedy policy is often used with Q-learning,
random exploration (with an exploration probability) in physical, continuous-action domains can be inefficient. To explore actions well in our building environment, we decided to sample actions from a Gaussian distribution with its mean at the greedy actions and with a small fixed standard deviation
throughout the training to encourage life-long exploration. We used a fixed standard deviation of 0.2 during training and zero during evaluation. This exploration strategy yielded a mildly better performance as compared to using an $\epsilon-$greedy policy with a fixed or linearly annealed exploration probability. The duration of each time (action) slot is 15 minutes. We achieved convergence of our reward function after 1000 episodes as explained in Section~\ref{convergence-section}.


\begin{figure}[t]
\centering
  \includegraphics[height=1.6in, width=3.0in]{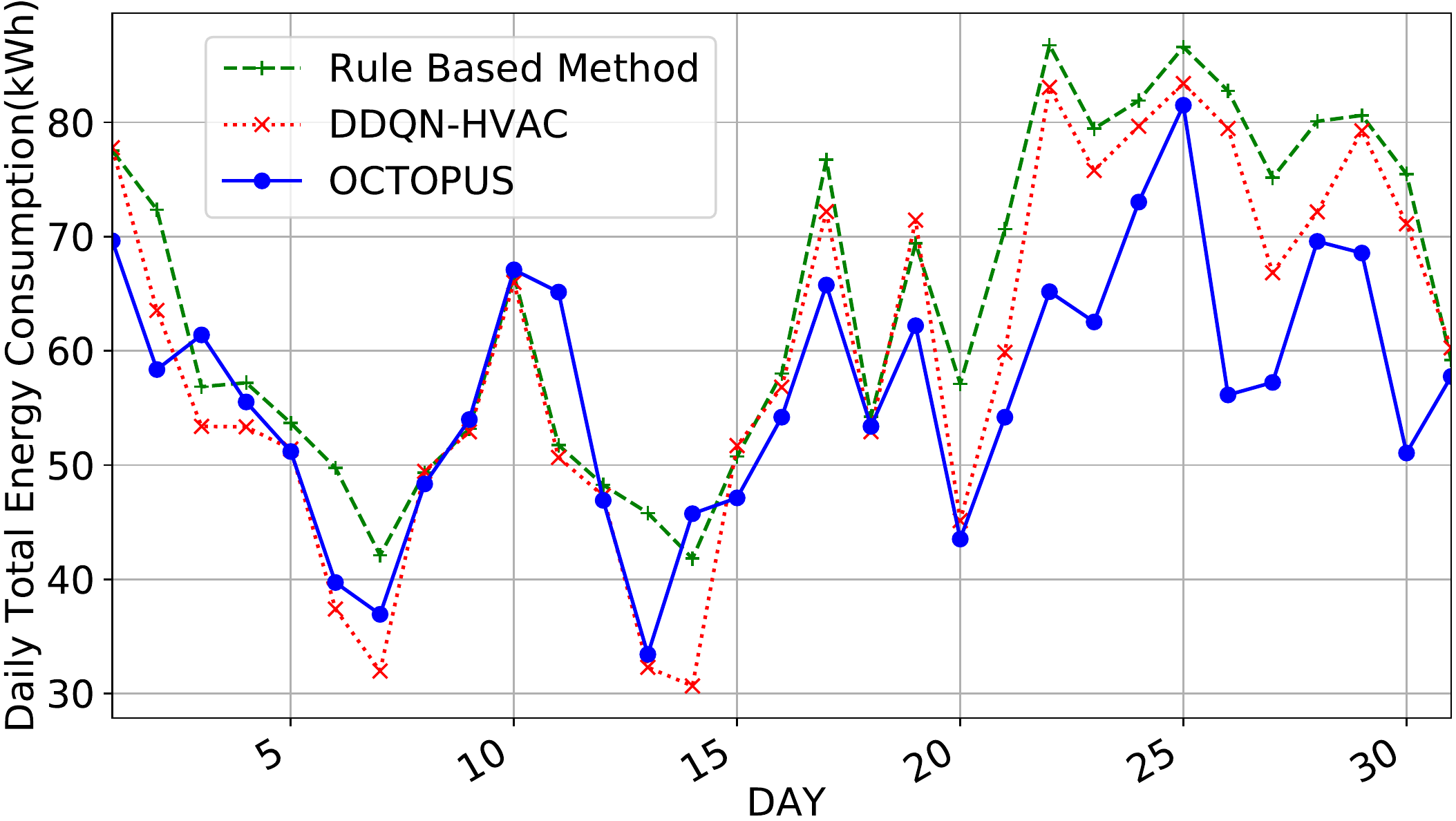}
  \caption{Daily Energy Consumption of Control Methods.}
  \label{energy_trend}
\end{figure}

\begin{figure}[t]
\centering
  \includegraphics[height=1.6in, width=3.0in]{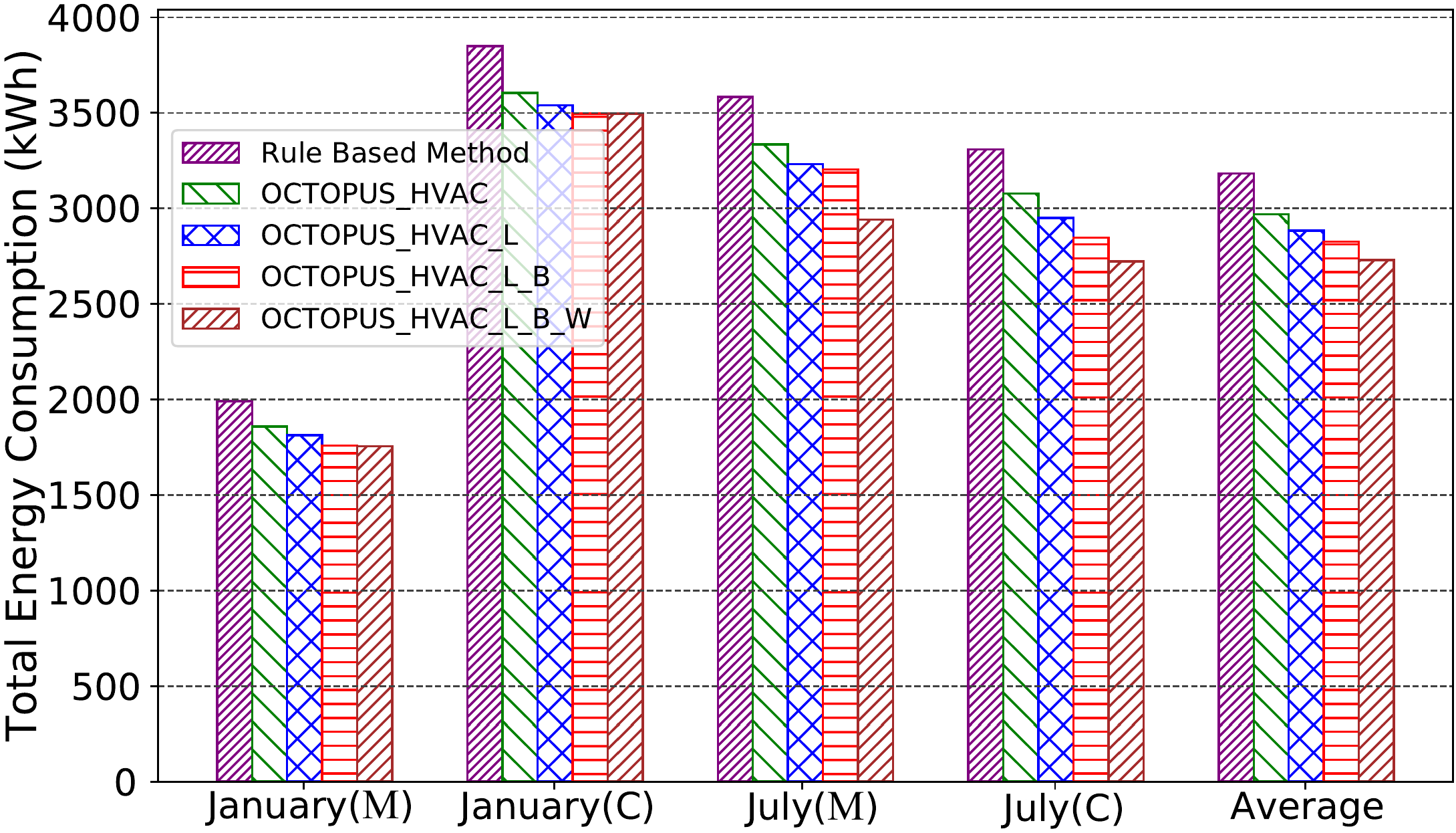}
  \caption{Performance Contribution of Each Subsystem.}
  \label{different_subsystem}
\end{figure}

\begin{figure}[t]
\centering
  \includegraphics[height=1.6in, width=3.0in]{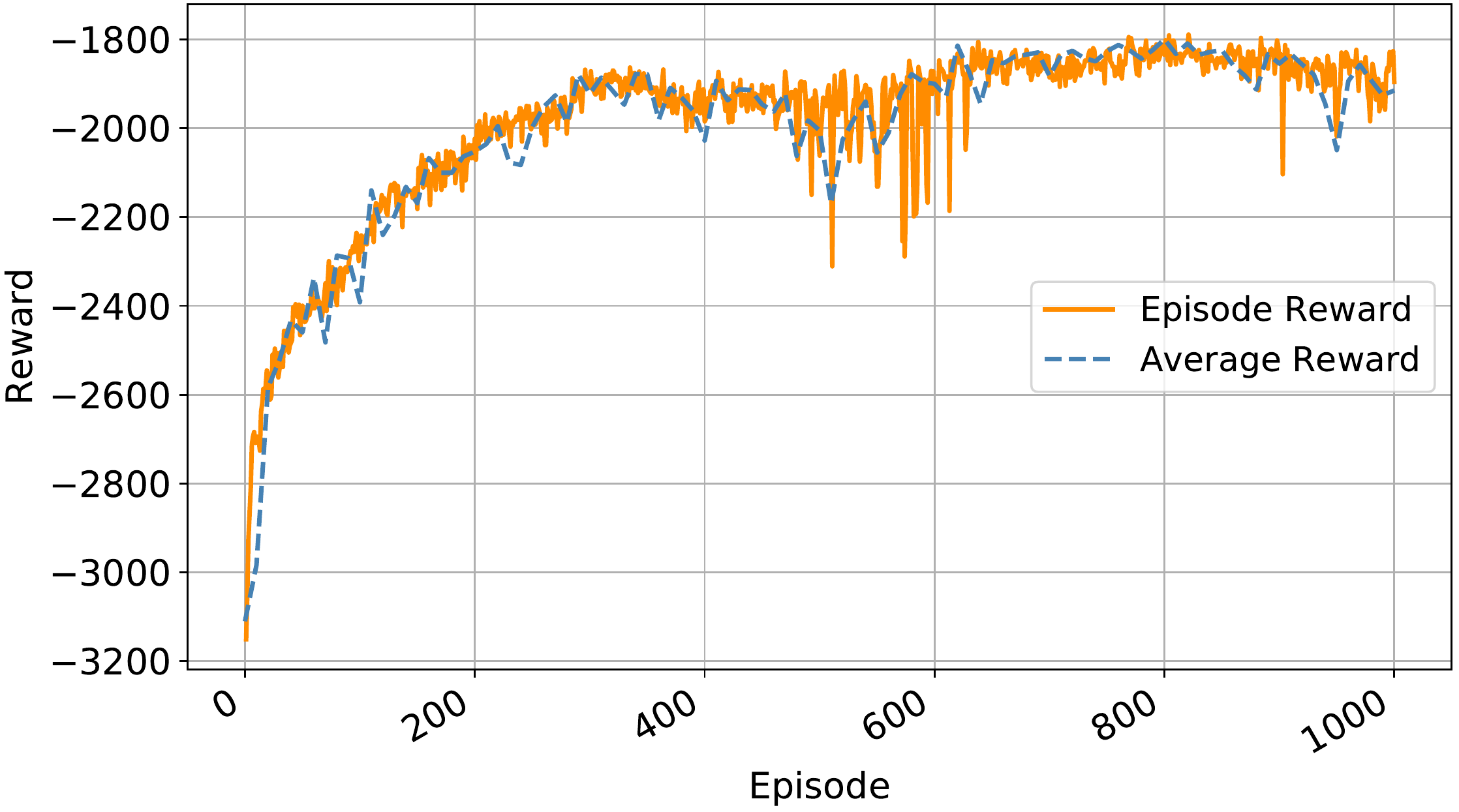}
  \caption{The Convergence of \aliasAPP.}
  \label{reward_convergence}
\end{figure}

\section{Evaluation}\label{sec:evaluation}
In this section, we compare the performance of \aliasAPP with the rule-based method and the latest DRL-based method.

\vspace{-0.05in}
\subsection{Experiment Setting}
The implementation of the rule-based HVAC control has been introduced in Section~\ref{rule_based_method}. 
The rule-based method only controls the HVAC system. 
For the conventional DRL-based method, we implement the dueling DQN architecture used in~\cite{zhang2018practical}, which controls the water-based heating system. We name that work as DDQN-HVAC in our comparison. 
Since these two benchmarks do not control the light system, for a fair comparison, we initialize the lights on in all experiments. \aliasAPP may dim the lights if the blind is open during the day.
In addition, the two benchmarks always leave the blind and window system closed.

The three human comfort metrics are measured by PMV, Illuminance, and carbon dioxide concentration. 
We set the acceptable range of three human comfort metrics according to building standards and previous experiences in related work.
The comfort range of PMV is set to -0.5 to 0.5~\cite{ashrae2017_thermal_comfort}. The comfort range of illuminance is set to 500-1000 lux~\cite{pritchard2014lighting}. The comfort range of carbon dioxide concentration is set to 400-1000 ppm~\cite{ashrae_indoor_air_condition}. 

We use three control methods to control the building we modeled in Section~\ref{sec:Implementation} for two months (January and July) and at two places with distinct weather patterns. 
Table~\ref{human_comfort} shows the human comfort results of three control methods and their energy consumption. 
The violation rate is calculated as the time when the value of a human comfort metric falls beyond its acceptable range divided by the total simulated time. Other quality of service metrics, including the amount by the which the violation occurred, or combination of amount and time will be explored in future work.

\subsection{Human Comfort} 
From the results in Table~\ref{human_comfort}, we see that all three methods can maintain the PMV value in the desired range for most of the time since the violation rate is low. 
The average PMV violation rate of \aliasAPP and DDQN-HVAC is higher than the rule-based method by 2.19\% and 2.22\% respectively. 
The reason for this is that the DRL-based methods try to save more energy by setting the PMV to a value close to the boundary of the acceptable range. It can be observed in Table~\ref{human_comfort} that the average PMV value of \aliasAPP and DDQN-HVAC (-0.36 and -0.26) is closer to the range boundary (-0.5), compared with the rule-based method (-0.13). 

For both visual comfort and indoor air quality, the three control methods provide a very small violation rate. For illuminance, the mean illuminance value of \aliasAPP and DDQN-HAVC is 590.69 lux and  610.89 lux respectively. \aliasAPP saves energy by utilizing natural light as much as possible. 
For indoor air quality, the average of $CO_{2}$ concentration of \aliasAPP, DDQN-HVAC, and rule-based method is 620.28 ppm, 633.92 ppm, and 635.06 ppm. \aliasAPP adjusts both window system and HVAC system to maintain the $CO_{2}$ concentration level within the desired range. DDQN-HVAC and the rule-based method only use the HVAC system.

\subsection{Energy Efficiency}
The results in Table~\ref{human_comfort} reveal that \aliasAPP save 14.26\% and 8.1\% energy on average, compared with the rule-based control method and DDQN-HVAC. In both cities, \aliasAPP achieves similar performance gain.
\aliasAPP reduces the energy consumption of HVAC by using the other subsystems. 
Figure~\ref{energy_trend} shows a daily energy consumption of three control methods in January at Merced. In most days, \aliasAPP consumes less energy than the other two methods; however, \aliasAPP is not always the best although we see clear gains towards the second half of the month due to a change in weather temperature. The average range of outdoor temperature changes from 2 $^{\circ}C$  $\sim$ 13 $^{\circ}$C in the first half of the month to -1 $^{\circ}$C $ \sim$ 18 $^{\circ}$C in the second half of the month. \aliasAPP could use external air with the window open for more natural ventilation.

In Table~\ref{human_comfort}, compared to the rule-based method and DDQN-HVAC, \aliasAPP saves more energy in July (17.6\% and 11.7\%) than in January (10.05\% and 3.9\%). 
In July, the outdoor air temperature range at Merced and Chicago is 15$^{\circ}$C $ \sim$ 42$^{\circ}$C and 15$^{\circ}$C $\sim$ 40$^{\circ}$C respectively. The window can be opened when the temperature is within the acceptable range, in order to save the energy consumed by the HVAC system.
However, in January, due to the cold weather at both places, the windows stay closed most of the time and cannot make much contribution to energy savings.

\subsection{Performance Decomposition}
We implement four versions of \aliasAPP to study the energy saving contribution of each subsystem, i.e., \aliasAPP just with the HVAC system (\aliasAPP\_HVAC), \aliasAPP with HVAC and lighting (\aliasAPP\_HVAC\_L), \aliasAPP with HVAC, lighting and blind (\aliasAPP\_HVAC\_L\_B) and \aliasAPP with all four subsystems (\aliasAPP\_HVAC\_L\_B\_W). 
Figure~\ref{different_subsystem} depicts the energy consumption of these four versions in two different months and at two different places (Merced and Chicago). 
Compared with the rule-based method, \aliasAPP\_HVAC can save 6.16\% more energy by only considering HVAC. When the lighting system is added in \aliasAPP\_HVAC\_L, 2.73\% more energy can be saved.
If the blind system is further added in \aliasAPP\_HVAC\_L\_B\,  
1.93\% more energy can be saved.
Finally, when the window system is added in \aliasAPP\_HVAC\_L\_B\_W, 3.44\% more energy can be saved.
Four subsystems make different contributions to energy saving in January and July. In January, four subsystems (i.e., HVAC, lighting, blind and window) make 6.16\%, 2.73\%, 1.93\% and 0\% contribution of energy savings respectively. In July, the contribution of these subsystems changes to 5.9 \%, 3.31 \%, 1.99\%, and 6.4\% respectively. The most obvious difference between these two months is made by the window system (6.4\%).
The reason for this has been explained above. In January, the windows are closed almost all the time. In July, the cold outdoor air is used to cool down the building instead of using HVAC system.

\begin{table*}[t]
  \renewcommand\arraystretch{1.2}
  \small
\caption{Different Parameters for Reward Function in Octopus}
\centering
\begin{tabular}{c|c|c|c|c|c|c|c}

\cline{1-8}

\shortstack{Parameter\\($\rho _{1}$,$\rho _{2}$, $\rho _{3}$, $\rho _{4}$)}
 & 
 \multicolumn{2}{c|}{PMV}
 
 & \multicolumn{2}{c|}{\shortstack{Illuminance\\(lux)} }& \multicolumn{2}{c|}{\shortstack{CO$_{2}$ Concentration\\(ppm)} }&\shortstack{Energy (kWh)} \\
 
\cline{2-7}
  & Mean &Std &Mean&Std &Mean&Std &\\
\cline{1-8}
  1, 1, 1, 1 & -0.36&0.15&587.35&94.52&587.25&101.14&3250.55 \\
\cline{1-8}
    5, 1, 1, 1&-0.33& 0.16 &   611.71& 131& 608.48&175.1&3221.20 \\
\cline{1-8}				
  10, 1, 1, 1 & -0.31& 0.16  &  624.97& 189.04& 	647.77&150.33&3150.62 \\
\cline{1-8}
  2, 3, 1, 1 & -0.383& 0.10  &  569.88& 75.83& 	636.5&179.46&2941.46 \\
\cline{1-8}
  2, 5, 1, 1 & -0.481& 0.13  &  689.23& 146.66& 	616.02&177.32&2900.44 \\
\cline{1-8}
\end{tabular}

\label{hyperparameters}
\end{table*}

\subsection{Hyperparameters Setting} 
The hyperparameters in the reward function (Equation \ref{lagrangian}) are
tuned to balance between the energy consumption and human comfort. Table \ref{hyperparameters} shows the performance results of the trained DRL agents in the selected experiments of the hyperparameters tuning. The total energy consumption and the mean and standard deviation of the PMV, Illuminance and carbon dioxide concentration are used as the evaluation metrics. It is interesting to find that the control performance results of the different hyperparameters are not intuitive. For example, we would expect the bigger $\rho _{1}$ and smaller $\rho _{2},\rho _{3},\rho _{4}$ to lead to lower energy consumption and just meet the requirements of thermal comfort, visual comfort and indoor air condition. However, the results in Table \ref{hyperparameters} shows that when increasing the weight of energy, energy consumption does not necessarily decrease. Such counter-intuitive results are possibly caused by the delayed reward problem that the DRL agents are stuck in some local optimal areas during    the training. Out of the five experiments in Table \ref{hyperparameters}, the fourth row saves 17.9\% of the energy consumption with only slightly worse three human comfort quality in the testing model, which comparably achieves the best balance between the human comfort and energy consumption. Therefore, the parameters in the fourth row are used for the trained agent.

\vspace{-0.05in}
\subsection{Convergence of \aliasAPP training} 
\label{convergence-section}
Figure \ref{reward_convergence} shows that the accumulated reward of \aliasAPP in each episode during a training process.
We calculate the reward function every control time step (15 minutes), and thus one episode (one month)  contains 2880 time steps. 
The accumulated reward of one episode (episode reward in Figure \ref{reward_convergence}) is the sum of the rewards of 2880 time steps.
From the results in Figure \ref{reward_convergence}, we see that the episode reward increases and tends to be stable as the number of training episodes increases. 
When the episode reward does not change much, it means that we cannot do further to improve the learned control policy and thus the training process converges. 
As indicated in Figure \ref{reward_convergence}, the training reward fluctuates between two adjacent episodes, because the number of time steps is large in one episode, i.e., 2880. 
The rewards calculated at some of these 2880 time steps may vary dynamically because we randomly choose some time steps by an exploration rate (determined by a Gaussian distribution with a standard deviation of 0.2). At these time steps, we do not use the action generated by the agent, but randomly choose an action to avoid local minimum convergence. 
If we smooth the episode reward using a sliding window of 10 episodes, the average reward in Figure \ref{reward_convergence} is more stable during the training.

\section{Discussion}

\textbf{Deploying in a Real Building.}
Although we have developed a calibrated simulation model of a real building on our campus for training and evaluation, we have not deployed \aliasAPP in the building, because we do not have access to automatic blind and window system at the moment. We are seeking financial support to work with our facility team for a possible upgrade.
\aliasAPP is designed for real deployment in buildings. For a new building, we need to build an EnergyPlus model for it and calibrate the model using real building operation data. After training the \aliasAPP control agent using the calibrated simulation model and real weather data, we can deploy the trained agent in the building for real-time control. For a certain action interval (e.g., every 10 mins), the \aliasAPP control agent takes the state of the building as input and generates the control actions of four subsystems. \aliasAPP can provide real-time control, as one inference only takes 22 ms. 
We plan to deploy \aliasAPP in a real building in our future work.  

\textbf{Scalability of \aliasAPP.} \aliasAPP can work in a one-zone building with one HVAC system, lighting zone, blind and window. However, a realistic building (or even a small home) is usually equipped with many lighting zones, blinds and windows which may take different actions in one subsystem. \aliasAPP may solve this scalability problem by increasing the number of BDQ branches, i.e., each branch corresponds to one subsystem in each zone of a building. We will tackle this scalability problem in our future work.

\textbf{Building Model Calibration.}
A critical component of our architecture is the use of a calibrated building model that is close to the target building, allowing us to generate sufficient data for our training needs. However, getting a calibrated model "right" is a tedious process of trial-and-error over a large number of parameters. Out of the thousands of parameters available in EnergyPlus, we use our experience and consulted experts to determine both the most important parameters and a sensible range of values to explore (it took us four weeks to get it "right"). However, there is no magic bullet, and this may become a problem, especially for unusual building architectures or specialized HVAC systems that may not be trivial to replicate in a simulation environment.

\textbf{Accepting Users' Feedback.}
Some existing work~\cite{winkler2016forces} allows users to send their feedback to the control server. The feedback can represent a user's personalized preference on different human comfort metrics and will be considered in the control decision process. 
\aliasAPP can easily accept users' feedback to train a better agent model by making a small modification, i.e., changing the calculated comfort values in the reward function by the users' feedback. 
This can be used for the initial training or for updated training (once deployed). For example, the \aliasAPP control agent can be trained incrementally with a certain time interval (e.g., one month). The newly-trained agent will be used for real-time.

\section{Conclusions}\label{sec:conclusion}
This paper proposes \aliasAPP, a DRL-based control system for buildings that holistically controls many subsystems in modern buildings (e.g., HVAC, light, blind, window) and manages the trade-offs between energy use and human comfort. As part of our architecture, we develop a system that addresses the issues of large action state, a novel reward function based on energy and comfort, and data requirements for training using existing historical weather data together with a calibrated simulator for the target building.  We compare our results with both the state-of-art rule-based control scheme obtained from a LEED Gold certified building, a DRL scheme used for optimized heating in the literature, and show that we can get 14.26\% and 8.1\% energy savings while maintaining (and sometime even improving) human comfort values for temperature, air quality and lighting.




\bibliographystyle{IEEEtran}
\bibliography{barejrnl}

\newpage

 




\vfill

\end{document}